\documentclass[10pt, conference]{IEEEtran}

\usepackage[style=acmnumeric]{biblatex}
\usepackage{amssymb}
\usepackage{amsmath}
\usepackage[colorlinks=true, allcolors=blue]{hyperref}
\usepackage{float}
\usepackage{bm}
\usepackage{graphicx}
\usepackage{booktabs}
\usepackage{tabularx}
\usepackage{subcaption}
\usepackage{dsfont}
\usepackage{authblk}
\usepackage{makecell}
\usepackage{placeins}

\newcommand{\vect}[1]{\bm{#1}}

\title{Probabilistic Wind Power Forecasting with Tree-Based Machine Learning and Weather Ensembles}

\author[1]{Max Bruninx}
\author[1,2]{Diederik van Binsbergen}
\author[1]{Timothy Verstraeten}
\author[1]{Ann Nowé}
\author[1]{Jan Helsen}
\affil[1]{Vrije Universiteit Brussel, Brussels, Belgium} 
\affil[2]{Norwegian University of Science and Technology, Trondheim, Norway}

\begin{document}
\maketitle

\begin{abstract}
    Accurate production forecasts are essential for the integration of renewable energy sources into the power grid. This paper illustrates how to obtain probabilistic forecasts of wind power generation using gradient boosting trees and an ensemble of weather forecasts. To this end, we perform a comparative analysis across three state-of-the-art probabilistic prediction methods—conformalized quantile regression, natural gradient boosting and conditional diffusion models—all of which can be combined with tree-based machine learning. The methods are validated using four years of data for all Belgian offshore wind farms. We benchmark the models against the power curve and a calibrated wake model as well as a probabilistic method using stochastic variational Gaussian process regression. The tree-based models significantly reduce the mean absolute error in comparison to the deterministic baselines. Additionally, all three methods outperform the Gaussian process baseline in probabilistic skill, while two out of the three also improve point forecast accuracy. The conditional diffusion model attains the best performance, with improvements of 5\% in mean absolute error and 12\% in continuous rank probability score compared to the probabilistic baseline. Last, the results indicate an average improvement in point forecast accuracy of 17\% by using an ensemble of weather forecasts instead of a single provider.
\end{abstract}

\section{Introduction}

Over the past two decades, the annual growth rate of renewable energy capacity has continued to increase, significantly reducing fossil fuel consumption and global emissions \cite{IEA_GlobalEnergyReview2025}. Nonetheless, the intermittent nature of renewable energy sources poses additional challenges and costs for the power system \cite{heptonstall2021systematic, weber2024intermittency}. Enhancing production forecasts is critical to mitigate these issues and facilitate the continued integration of renewable energy sources into the grid. The importance of this subject is also highlighted by the extensive body of research produced over the past fifty years \cite{HONG2016896}. Depending on the operational context, these forecasts may require different look-ahead horizons or granularity. Day-ahead forecasts, for example, can be used to bid on the day-ahead market \cite{pinson2007trading, heiser2024betting, bruninx2025day} or to determine a unit commitment schedule \cite{hou2021data, hou2022hybrid}. Intra-hour forecasts, on the other hand, can help maintain grid stability in real-time \cite{zhu2012short, ahmadi2025enhancing}. Furthermore, these forecasts can be either point forecasts, which estimate the conditional mean, or probabilistic forecasts, which capture the entire conditional probability distribution. The widespread recognition in the literature that weather variables should be modelled as stochastic processes (see, for example, \cite{allen2002towards, pinson2013wind}) has shifted the field towards the latter approach \cite{sweeney2020future}.

\begin{figure}
    \centering    \includegraphics[width=0.8\linewidth]{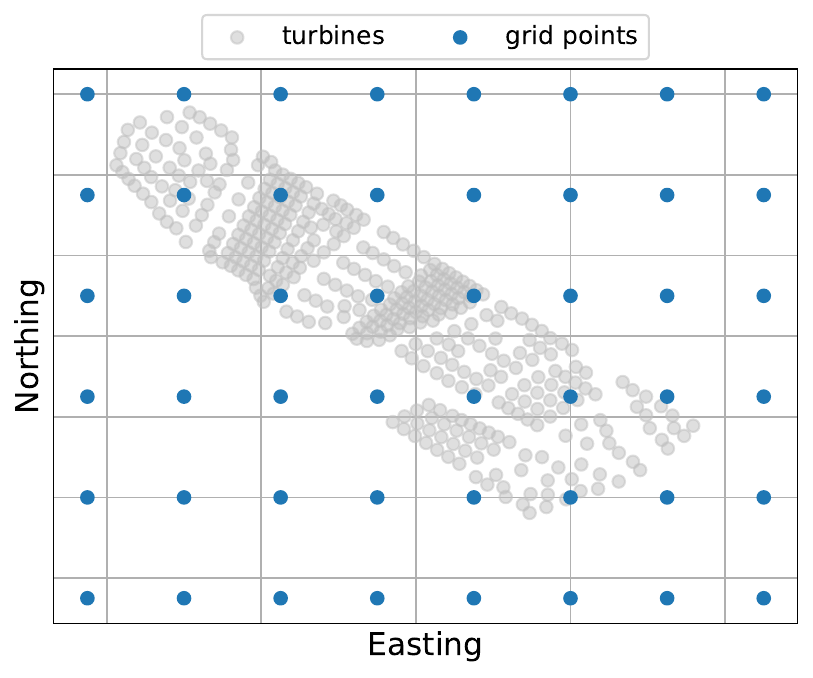}
    \caption{Location of the turbines in the Belgian offshore zone and the grid points from the DWD ICON-EU model (with a horizontal grid spacing of $6.5$ km).}
    \label{fig:nwp_grid}
\end{figure}

This work studies probabilistic day-ahead forecasting of wind power generation in the context of offshore wind farms. Nonetheless, the presented methodology can also be applied to different renewable energy sources or forecast horizons. Typically, Numerical Weather Predictions (NWPs) are the most important input for day-ahead forecasting, and statistical or machine learning models are employed to learn the relationship between weather forecasts and power output \cite{giebel2017wind}. The primary advantage of data-driven techniques lies in their ability to adapt wind forecasts to site-specific conditions and mitigate meteorological errors. This is particularly important in the context of day-ahead forecasting, since weather forecasts typically present a high degree of uncertainty at this stage. More specifically, the use of meteorological data introduces two sources of error relative to the true on-site wind conditions. First, the weather forecast error, representing the difference between the wind forecasts and the corresponding observations, and second, the location error, which stems from the spatial mismatch between the numerical weather prediction grid and the precise turbine locations. In typical NWP models, grid points are 7-12 km apart, whereas the spacing between offshore wind turbines is around 1 km, depending on the rotor size and the farm layout \cite{corke2018wind}. This is also illustrated in Figure \ref{fig:nwp_grid} for the Belgian offshore zone. Moreover, the local wind speeds in the wind farm are affected by wake effects \cite{gonzalez2012wake} and blockage \cite{segalini2020blockage}, which are not accounted for by the NWP models. In addition to the errors introduced by employing meteorological data, power production can also be influenced by other atmospheric variables, such as turbulence intensity, shear and veer \cite{he2026short}, along with turbine-specific characteristics.

In our work, we will consider tree-based machine learning methods, in particular gradient boosting trees, due to their superior performance in recent wind power forecasting competitions \cite{HONG2016896,bellinguer2020probabilistic,browell2025hybrid}. Furthermore, gradient boosting trees are often found to outperform deep learning methods on tabular datasets \cite{borisov2022deep, grinsztajn2022tree} and are generally more interpretable. However, these models produce discrete outputs by default and are therefore not directly applicable to probabilistic forecasting. To address this limitation, they can be integrated with probabilistic prediction methods to transform their output into predictive distributions. Probabilistic prediction methods can be categorized as either parametric methods, which assume an underlying parametric probability distribution, or non-parametric methods, also referred to as distribution-free, which do not make any assumptions on the underlying distribution of the target variable. 

Quantile regression \cite{koenker1978regression}, a non-parametric method, has emerged as the leading approach in the field of renewable energy \cite{sweeney2020future}. This method is model-agnostic, as it can be applied to any model that minimizes a loss function, such as gradient boosting trees \cite{mason1999boosting}. Conformalized quantile regression \cite{romano2019conformalized} extends upon this approach by calibrating the prediction intervals to ensure that they achieve statistically valid coverage. Treeffuser \cite{beltran2024treeffuser}, another non-parametric method inspired by recent advances in generative modelling, applies conditional diffusion models on tabular data using gradient boosting trees. This method produces probabilistic forecasts by drawing samples from the generative distribution learned by the model. Empirically, Treeffuser was found to achieve superior performance on a wide range of publicly available datasets from other fields. In terms of parametric methods, natural gradient boosting \cite{duan2020ngboost} and probabilistic gradient boosting machines \cite{sprangers2021probabilistic} are the most well-known in the area of tree-based machine learning. Whereas the former can estimate the parameters of any probability distribution, the latter is limited to distributions for which the parameters can be expressed in terms of mean and variance.

This study aims to assess which probabilistic prediction methods are the most suitable in the context of wind power forecasting. For this purpose, we consider three state-of-the-art methods, namely conformalized quantile regression, natural gradient boosting and conditional diffusion models.  The main contributions of the article are the following:
\begin{itemize}
    \item We demonstrate how these three methods can be combined with gradient boosting trees to generate probabilistic forecasts of wind power production. Although the first two methods have been applied in prior research on wind power forecasting \cite{hu2022conformalized,wang2023conformal,zhang2024two, li2020short}, this study is the first to directly compare them and to include the third approach. 
    \item We provide an extensive validation of the different methods, considering four years of data and encompassing all wind farms within the Belgian offshore zone. This contrasts with existing literature, which often relies on datasets restricted to one or two wind farms and shorter time horizons. Furthermore, we consider an ensemble of weather forecasts rather than a single weather forecast and demonstrate how this improves predictive performance.
    \item In addition to a machine learning baseline, we also benchmark the methods against deterministic engineering approaches, employing a well-established method using the power curve and an advanced approach based on a calibrated analytical wake model. This direct comparison against engineering methods is often overlooked in existing studies using machine learning. 
\end{itemize}

The remainder of the paper is organized as follows: Section \ref{sec:methodology} provides an overview of the forecasting methodology, covering the different methods as well as the use of an ensemble of weather forecasts. Section \ref{sec:experiments} details the experiments conducted across all wind farms within the Belgian offshore zone, while Section \ref{sec:results} covers the experimental outcomes. Finally, Section \ref{sec:conclusion} concludes the work.

\textit{Note}: Throughout the paper, symbols with a hat $\hat{(.)}$ denote model predictions and bold symbols indicate vectors. For an observation $i$, the target variable is given by $y_i$ and the feature vector by $\vect{x}_i$. $F_i(.)$ is the cumulative density function and $q_{i,\tau}$ or $F^{-1}_i(\tau)$ are used interchangeably for the $\tau$-quantile prediction. Last, the significance level $\alpha\in[0,1]$, determines the $(1-\alpha)\%$ confidence interval $\mathcal{C}_\alpha$.

\section{Methodology}
\label{sec:methodology}

\begin{figure*}[ht!]
    \centering
    \includegraphics[width=0.9\textwidth]{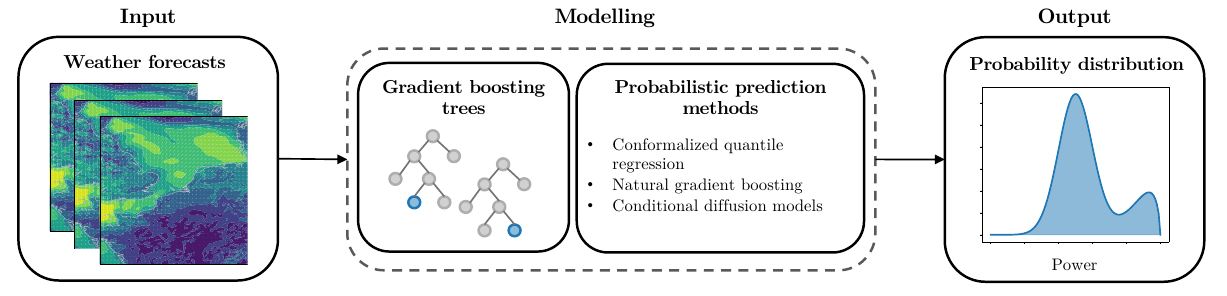}
    \caption{High-level overview of the probabilistic forecasting methodology. The framework integrates three different probabilistic prediction methods with gradient boosting trees to model the conditional power distribution. The method employs different day-ahead weather forecasts as input and aims to estimate the distribution of the power output of a given wind farm.}
    \label{fig:overview}
\end{figure*}

Figure \ref{fig:overview} presents an overview of the probabilistic day-ahead forecasting methodology. The method aims to model the conditional power distribution given an ensemble of different weather forecasts. To this end, gradient boosting trees are combined with three state-of-the-art probabilistic prediction methods: conformalized quantile regression, natural gradient boosting and conditional diffusion models. Section \ref{sec:methodology_input} discusses the different weather forecasts and how they are preprocessed. Sections \ref{sec:methodology_cqr} to \ref{sec:methodology_cd} cover the details of the different probabilistic prediction methods. Last, Section \ref{sec:methodology_benchmark} explains the deterministic engineering models used to evaluate point forecast performance.

\subsection{Weather forecasts}
\label{sec:methodology_input}

Following the typical model chain, the day-ahead wind speed and wind direction forecasts are used as input to the models \cite{giebel2017wind}. Instead of relying on one weather forecast provider, we combine forecasts from different providers to incorporate the uncertainty among these forecasts. The forecasts are further processed by including lagged variables, i.e., the forecasts of the previous and the next hours, similar to the approach in \cite{ally2025modular}. Furthermore, summary statistics such as the mean and standard deviation are calculated for each hour across the different providers. 

Due to the comparative nature of this study, weather forecasts were restricted to a single representative grid point located at the center of the Belgian offshore zone. This approach allows us to compare the different methods for each farm using the same input. However, the chosen grid point could have a negative impact on the predictive accuracy of wind farms located at the periphery of the offshore zone. It is therefore important to highlight that we do not aim to develop the most accurate forecast for an individual farm. Instead we want to compare the relative performance of the forecasting models under various operating conditions and strategies. Nonetheless, empirical results have suggested that considering multiple grid points could improve forecast accuracy \cite{andrade2017improving}, and therefore extending the methodology to a spatio-temporal context remains an interesting direction for future work.

\subsection{Conformalized quantile regression}
\label{sec:methodology_cqr}

In a quantile regression model \cite{koenker1978regression}, the quantile loss
\begin{equation}
    L(y_i,\hat{q}_{i,\tau}) = \begin{cases}
        \tau\left(y_i-\hat{q}_{i,\tau}\right) & y_i\geq\hat{q}_{i,\tau}\\
        (1-\tau)\left(\hat{q}_{i,\tau}-y_i\right) & y_i<\hat{q}_{i,\tau}
    \end{cases},
\label{eq:quantile_loss}
\end{equation}
is used to directly estimate the $\tau$-quantile $q_{i,\tau}$ rather than the target variable. This method is flexible, as it can be applied to any model that minimizes a loss function, and distribution-free, since it does not assume an underlying distribution of the target variable. However, it should be noted that the model provides no guarantee regarding the coverage of the intervals defined by the quantile predictions. To address this limitation, \cite{romano2019conformalized} proposed to combine this method with conformal prediction.

Conformal prediction is a statistical calibration procedure which can be used to attain a marginal coverage guarantee for a prediction set:
\begin{equation}
    1-\alpha \leq \mathbb{P}\left(y_i\in \mathcal{C}^{\mathrm{CP}}\left(\vect{x}_i\right)\right) \leq 1 - \alpha + \frac{1}{n+1}, 
\end{equation}
with $(\vect{x}_i, y_i)$ an unseen observation from the same data distribution as the calibration set, $\mathcal{C}^{\mathrm{CP}}\left(\vect{x}_i\right)$ the prediction set after calibration and $n$ the number of samples in the calibration set. To this end, the conformal score is first computed for each observation in a hold-out calibration set. Afterwards, the $\frac{(n+1)(1-\alpha)}{n}$ empirical quantile of this score is used to adapt the prediction set. The conformal score quantifies how the observed value differs from the model output and is also referred to as a nonconformity measure. An overview of conformal scores in a regression setting can be found in \cite{kato2023review}. We also refer the reader to \cite{angelopoulos2021gentle} for an introduction to conformal prediction. 

\begin{figure}[t!]
    \centering
    \includegraphics[width=0.8\linewidth]{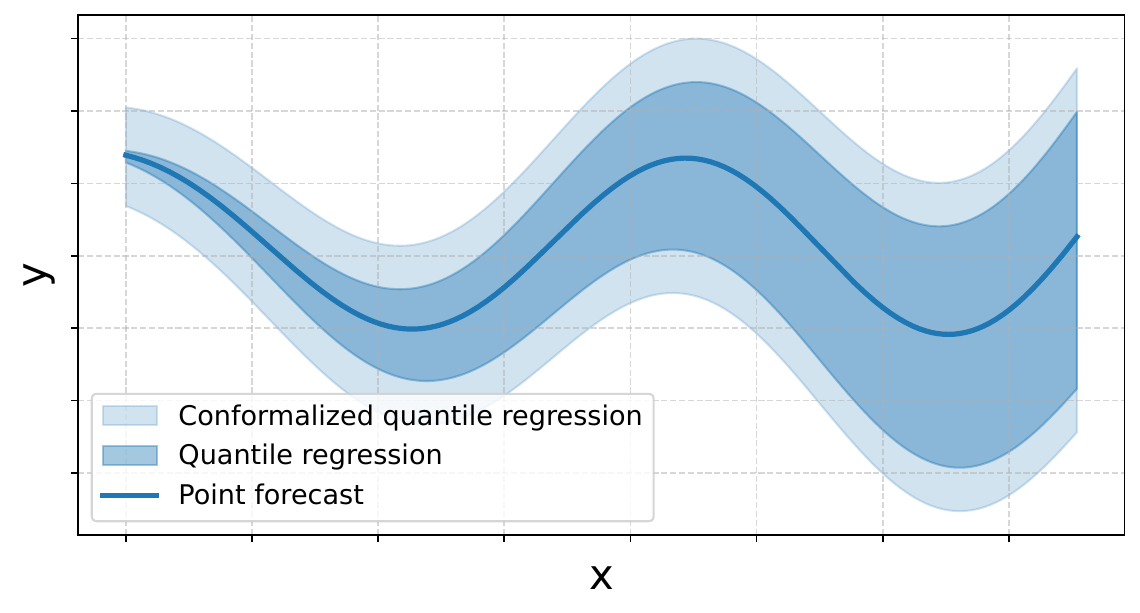}
    \caption{Illustrative example of the conformal prediction procedure. The dark blue area shows the confidence interval predicted by the quantile regression model, while the light blue area illustrates how this interval changes after conformal calibration. In this example, the empirical quantile of the conformal scores, denoted by $\hat{s}$, is positive. This indicates that the initial prediction interval did not provide sufficient coverage on the calibration set and had to be expanded.}
    \label{fig:cqr}
\end{figure}

In the context of conformalized quantile regression, the conformal score for the confidence interval obtained by quantile regression
\begin{equation}
    \mathcal{C}_\alpha(\vect{x}_i) = \left[\hat{q}_{i,\alpha/2},\ \hat{q}_{i,1-\alpha/2}\right],
\end{equation}
is given by:
\begin{equation}
    s(\vect{x}_i, y_i) = \max\left\{\hat{q}_{i,\alpha/2} - y_i,\ y_i - \hat{q}_{i,1-\alpha/2}\right\}.
    \label{eq:ci}
\end{equation}
The score is positive when the observed value lies outside of the prediction interval and increases when the distance towards the prediction interval increases. After computing the empirical quantile of the conformal scores on the calibration set, which we denote by $\hat{s}$, the confidence interval is adapted accordingly:
\begin{equation}
    \mathcal{C}^{\mathrm{CP}}_\alpha\left(\vect{x}_i\right) = \left[\hat{q}_{i,\alpha/2} - \hat{s},\ \hat{q}_{i,1-\alpha/2} + \hat{s}\right].
    \label{eq:cp_cqr}
\end{equation}
This is also illustrated visually in Figure \ref{fig:cqr}.

Note that quantile regression does not guarantee that the predicted quantiles are non-crossing, i.e., that lower-level quantiles are always lower or equal to higher-level quantiles. To resolve this issue, we employ the post-processing technique introduced by \cite{chernozhukov2010quantile}, which sorts the quantile predictions \textit{a posteriori} to enforce monotonicity. 

\subsection{Natural gradient boosting}
\label{sec:methodology_ngb}

Natural gradient boosting \cite{duan2020ngboost} is a parametric probabilistic prediction method that leverages gradient boosting. Unlike traditional gradient boosting regression, which only estimates the expected value, the model learns the parameters of a probability distribution that represents the target conditional on the input variables. In this work, we will assume a Normal or Gaussian distribution, with the cumulative density function given by:
\begin{equation}
    F(y; \mu,\sigma) = \frac{1}{\sigma\sqrt{2\pi}}\int_{-\infty}^{y} \exp\left(-\frac{(t - \mu)^2}{2\sigma^2}\right) dt.
\end{equation}
The conditional parameters, denoted by $\mu_i$ and $\sigma_i$, are simultaneously estimated by minimzing the scoring rule. To this end, we consider the logarithmic score:
\begin{equation}
    \mathcal{L}(y_i, \hat{\theta}_i) = -\log P_{\hat{\theta}_i}(y_i),
\end{equation}
with $\hat{\theta}_i = \left\{\hat{\mu}_i, \hat{\sigma}_i\right\}$ and $P_\theta$ the likelihood function. The scoring rule is optimized via gradient decent using the natural gradient. In contrast to the ordinary gradient, the natural gradient makes updates invariant to the choice of parameterization and leads to more stable learning dynamics.

\subsection{Conditional diffusion models}
\label{sec:methodology_cd}

Treeffuser \cite{beltran2024treeffuser} is a score-based conditional diffusion model for tabular data via gradient boosting machines. In score-based diffusion models \cite{song2021scorebased}, the forward diffusion process maps the data distribution $p_0$ to a noise distribution $p_T$ using a stochastic differential equation:
\begin{equation}
    d\vect{y} = f(\vect{y},t)dt +g(t)d\vect{w},
\end{equation}
with $\vect{w}(t)$ a standard Brownian motion, $f$ the drift function and $g$ the diffusion function. The reverse diffusion process, that can be used to generate samples from $p_0$ using $p_t$, is then given by:
\begin{equation}
    d\vect{y} = \left[ f(\vect{y}, t) - g^2(t) \nabla_{\vect{y}} \log p_t(\vect{y}) \right] dt + g(t) d\tilde{\vect{w}},
\end{equation}
where $\tilde{\vect{w}}$ is a standard Brownian motion with reverse time and $\nabla_{\vect{y}} \log p_t(\vect{y})$ represent the score function, which is unknown and should be learned from data by a model. Conditional diffusion models extend this to construct a diffusion model conditional on some input variables. This can either be attained by post-processing the output of a diffusion model using guidance methods \cite{dhariwal2021diffusion,ho2022classifier} or incorporating the information in the training step \cite{batzolis2021conditional, rombach2022high}. Treeffuser employs the latter approach. After training the model, probabilistic predictions can be obtained by sampling from the conditional generative distribution and computing the quantiles from these samples. In this work, we draw 50 samples to calculate the predicted quantiles.

\subsection{Deterministic benchmark models}
\label{sec:methodology_benchmark}

\subsubsection{Power curve}

The simplest way to obtain a deterministic power forecast, which is often used in practice, is to apply the wind speed forecast to the warranted power curve for each turbine present in the wind farm: 
\begin{equation}
    \hat{y}_i = \sum_{j=1}^{m} P^{\textrm{M}}_j(v_{i,s}),
\label{eq:power_curve}    
\end{equation}
with $m$ the number of turbines in the farm, $P^{\textrm{M}}_j(.)$ the power curve of turbine $j$ and $v_{i,s}$ the wind speed forecast. However, this method does not consider the wake effect present in wind farms, where upstream wind turbines affect the available wind energy of downstream turbines, resulting in reduced power production \cite{gonzalez2012wake}. Consequently, power forecasts produced using this method will overestimate the actual possible power production.

\subsubsection{Analytical wake models}

To account for wake effects, analytical wake models are commonly used. These steady-state models account for intra-farm and inter-farm wake effects by considering, respectively, the layout of the target wind farm as well as the neighbouring wind farms. To this end, the self-similar Gaussian wake model of Niayifar \& Porté-Agel \cite{niayifar2015new} as implemented in the PyWake framework \cite{Pedersen2023} is employed, where the deterministic power forecast is calculated as follows: 

\begin{equation}
    \hat{y}_i = \sum_{j=1}^{m} P^{\textrm{W}}_j(v_{i,s},v_{i,d}),
\label{eq:wake_model}
\end{equation}
with \(m\) being the number of turbines in the farm, \(P^{\textrm{W}}_j(.)\) the power production of turbine \(j\) calculated using the power curve and the local wind speed acquired from the analytical wake model, and \(v_{i,s}\), \(v_{i,d}\) are the wind speed and wind direction forecast, respectively. 

In the self-similar model, the wake recovery rate depends on two parameters: one that varies with the local turbulence intensity and one that remains constant. These two parameters are calibrated for the Belgian-Dutch offshore cluster using SCADA data from multiple wind farms, following the procedure described in \cite{Binsbergen2024} and applied in \cite{Binsbergen2024b,van2024performance}. The local turbulence intensity is obtained by combining the freestream turbulence intensity, which is calculated according to the IEC standard \cite{IEC61400-1}, and the wake-added turbulence intensity, resulting from the Crespo-Hernández model \cite{Crespo1996}. The velocity deficit due to wake overlap is calculated using linear superposition \cite{Katic1987}, consistent with \cite{niayifar2015new}. 

\section{Experiments}
\label{sec:experiments}

\subsection{Set-up}

\begin{figure}[t!]
    \centering
    \includegraphics[width=0.9\linewidth]{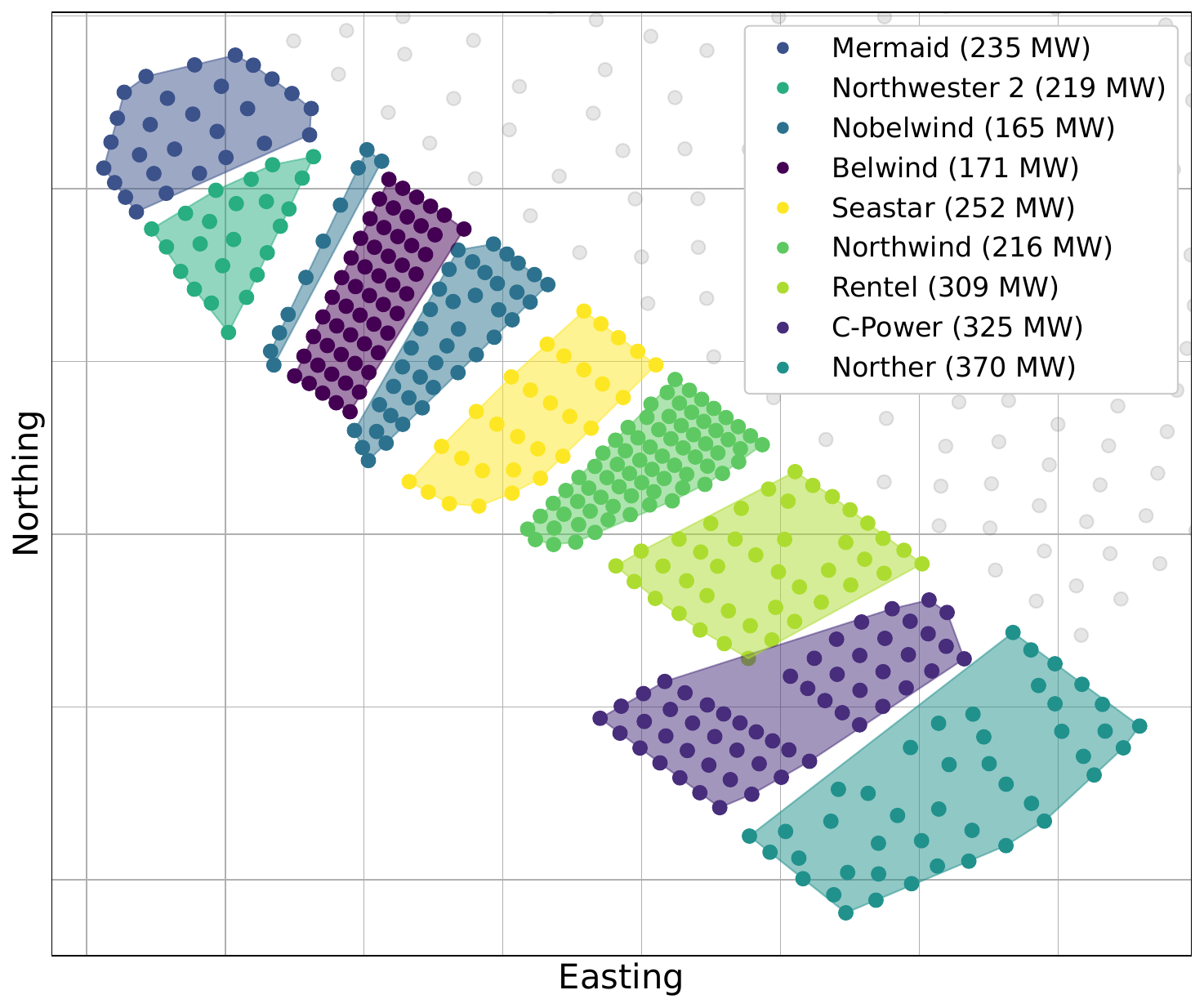}
    \caption{Overview of the wind farms present in the Belgian offshore zone (in color) and the surrounding wind turbines from the Borssele wind farm zone (NL) (in grey). The Belgian offshore zone has a total installed capacity of over 2 GW, divided among nine different wind farms and a total of 399 turbines. The first project, C-Power, was completed in 2009, while the most recent projects, Mermaid and Northwester 2, were finished in 2020. The Borssele wind farm zone has been fully operational since 2021.}
    \label{fig:offshore_zone}
\end{figure}

We evaluate the models on all wind farms present in the Belgian offshore zone, as visualized in Figure \ref{fig:offshore_zone}. To this end, we gather a dataset ranging from 2021 to 2024 with a resolution of one hour. The first three years are used to train the models, whereas the last year is used to evaluate the models. When a validation or calibration set is required, the data of 2023 is held-out from the training set. The hyperparameters of each model were optimized using random search with 25 iterations. The search space for hyperparameter tuning of the models can be found in \ref{app:hyp_opt}. For the Treeffuser model, we also include a model with the default hyperparameters since the authors suggested that tuning is not required \cite{beltran2024treeffuser}. The experiments were conducted on an Apple machine with an 8-core M3 chip and 24 GB of RAM.

In the remainder of the paper, the aforementioned models will be referred to as: Power curve (\ref{eq:power_curve}), Wake model (\ref{eq:wake_model}), CQR (Section \ref{sec:methodology_cqr}), NGBoost (Gaussian) (Section \ref{sec:methodology_ngb}), and Treeffuser (Section \ref{sec:methodology_cd}), along with its variant without hyperparameter tuning (Treeffuser (no tuning)). As a competitive probabilistic forecasting baseline, we have also implemented stochastic variational Gaussian process regression (SVGP) \cite{10.5555/3023638.3023667,hensman2015scalable}, with an ARD-RBF kernel and the inducing points initialized using k-means clustering.

\subsection{Data}

\subsubsection{Weather data}

The analysis incorporates five different weather forecasts, including the ICON-EU and ICON-D2 models from Deutscher Wetterdienst, as well as forecasts from the ECMWF HRES, Météo France ARPEGE-EU and Met Office Global Hi-Res models. These forecasts were obtained through a commercial weather forecast provider. For each day, the latest published forecast is selected before the day-ahead market gate closure time\footnote{In Belgium, the market gate closure time is 12:00 CET.}.  The weather forecasts are preprocessed as described in Section \ref{sec:methodology_input}. For the engineering models, the average day-ahead wind speed and direction forecasts over the different providers are used as input. Moreover, we also use ERA-5 reanalysis data obtained from the Open-Meteo API \cite{Zippenfenig_Open-Meteo}.

\subsubsection{Generation data}

Generation data at grid connection is obtained from the ENTSO-E Transparency Platform \cite{entsoe} for all wind farms in the Belgian offshore zone. Since the engineering methods do not account for grid losses present in the dataset, their predictions are rescaled using the ratio of the maximum observed power over the rated power. Moreover, we filter out curtailments from the training data to prevent the forecasts from showing a downward bias relative to the available wind resource. For example, if a wind farm operating at rated power receives a setpoint to reduce production by $50\%$, the resulting power output is no longer representative of the potential power generation. To this end, we remove observations involving activated downward balancing bids in the Belgian offshore zone using Elia Open Data \cite{eliaopendata}, reducing the size of the original dataset by $2.3\%$. Unfortunately, there is no information available on curtailments ordered by the Balance Responsible Party of the wind farm. As a proxy for these economic curtailments, we use ERA-5 reanalysis data to group observations into wind speed bins and remove any observation from the training set with a power below the 5th percentile of its bin. The bin size is set to 0.5 m/s, in accordance with the IEC standard \cite{IEC61400-1}, except when a bin contains fewer than 100 observations, in which case the bin size is incremented by 0.5 m/s until the bin contains sufficient observations. Depending on the wind farm, this procedure reduced the size of the original dataset by a value between $3.1\%$ and $3.6\%$. The 5th percentile threshold was chosen to identify the most important curtailments without discarding a substantial part of the dataset. For a threshold equal to the 1st or 10th percentile, the procedure would have resulted in removing around, respectively, $0.5\%$ or $7\%$ of the data.

\subsection{Performance metrics}

The models are evaluated on their point forecast and probability estimate. When evaluating the probability estimates, the 5th, 10th, 25th, 50th, 75th, 90th, and 95th percentiles of the conditional distribution are considered to ensure comparability between the methods. We define the point forecast of the models as the median prediction. The evaluation metrics are normalized by the installed capacity of the farm. On average the installed capacity per wind farm in the Belgian offshore zone is around 250 MW, implying that a relative forecast error of $1\%$ represents an absolute error of around $2.5$ MW.

\subsubsection{Mean absolute error}
The Mean Absolute Error (MAE) is used to evaluate the point forecasts of the models:
\begin{equation}
    MAE = \frac{1}{n}\sum_{i=1}^{n}|y_i-\hat{y}_i|,
\end{equation}
where $n$ equals the number of samples in the evaluation set.

\subsubsection{Continuous rank probability score}
Probability estimates are evaluated using the Continuous Rank Probability Score (CRPS) \cite{matheson1976scoring,hersbach2000decomposition}:
\begin{equation}
    CRPS =  \int_{-\infty}^{+\infty} \left(\hat{F}_i\left(t\right)-\mathds{1}\left(y_i\leq t\right)\ \right)^2\,dt,
\label{eq:crps}
\end{equation}
with $\mathds{1}\left(\cdot\right)$ the indicator function. This metric is then averaged over all observations. CRPS is a proper scoring rule, which measures reliability, i.e., how well the forecast probabilities align with the observed frequencies of events, and sharpness, i.e., how concentrated the probabilistic forecasts are around the actual outcomes \cite{hersbach2000decomposition, gneiting2007probabilistic}. Another interesting property is that for a point forecast the CRPS equals the MAE. Given that not all forecasting methods predict the entire cumulative density function, we employ the formulation proposed by \cite{nowotarski2018recent}:
\begin{equation}
    CRPS = \int_{0}^{1}L\left(y_i,\hat{q}_{i,\tau}\right)d\tau,
\end{equation}
where $L(.)$ equals the quantile loss (\ref{eq:quantile_loss}). For a discrete set of quantile predictions, the CRPS can be approximated by a Riemann sum.

\section{Results and discussion}
\label{sec:results}

\begin{table*}[ht!]
\centering
\caption{Out-of-sample normalized mean absolute error per wind farm, as well as the average score over the wind farms. The metric is normalized by the installed capacity of the wind farm. Lower scores indicate a better fit, with the best scores highlighted in bold.}\label{table:mae}
\small 
\setlength{\tabcolsep}{3.5pt} 
\resizebox{0.6\textwidth}{!}{
\begin{tabular}{lcccccccc} 
\toprule
 & Power & Wake & SVGP & NGBoost & CQR & \multicolumn{2}{c}{Treeffuser} \\
 & curve & model & & (Gaussian) & & (no tuning) & (tuned)\\
\midrule
Belwind       & 15.6\% & 10.5\% & 7.4\% & 7.5\% & 7.2\% & 7.1\% & \textbf{7.0\%} \\
C-Power       & 15.3\% & 10.8\% & 7.7\% & 7.7\% & 7.3\% & 7.1\% & \textbf{7.0\%} \\
Mermaid       & 14.5\% & 12.2\% & \textbf{8.9\%} & 9.2\% & 9.0\% & 9.2\% & 9.0\% \\
Nobelwind     & 17.6\% & 12.2\% & 8.5\% & 8.4\% & 8.3\% & 8.2\% & \textbf{8.1\%} \\
Norther       & 14.3\% & 10.9\% & 8.7\% & 8.5\% & 8.2\% & 8.1\% & \textbf{7.9\%} \\
Northwester 2 & 20.7\% & 15.2\% & \textbf{8.3\%} & 9.3\% & 8.9\% & 8.9\% & 8.9\% \\
Northwind     & 21.1\% & 11.8\% & 8.2\% & 8.2\% & 7.9\% & 7.7\% & \textbf{7.6\%} \\
Rentel        & 16.8\% & 11.8\% & 9.0\% & 8.8\% & 8.7\% & 8.7\% & \textbf{8.5\%} \\
Seastar       & 15.8\% & 11.9\% & 8.7\% & 8.8\% & 8.5\% & \textbf{8.4\%} & \textbf{8.4\%} \\
\midrule
Average       & 16.9\% & 11.9\% & 8.4\% & 8.5\% & 8.2\% & 8.2\% & \textbf{8.0\%} \\
\bottomrule
\end{tabular}
}
\end{table*}
Table \ref{table:mae} shows the mean absolute error of the different models on the test set. We observe that the distribution-free models outperform all of the baselines, but that the natural gradient boosting model achieves a slightly worse performance compared to SVGP. The Treeffuser model with parameter tuning is found to consistently achieve the best performance for all wind farms, with an improvement in MAE of 53\% compared to the power curve, 33\% against the wake model and 5\% against the probabilistic baseline. We also note that the analytical wake model improves MAE by 30\% compared to the power curve, indicating the importance of accurately modelling wake effects. Due to this significant difference in predictive performance, results derived from the power curve method will not be further discussed in the remainder of this paper. The improved performance of the data-driven methods against the calibrated wake model may be explained by the fact that the former can also account for the weather forecast error, whereas the wake model treats the wind speed and direction forecast as an exact input. In addition, we note that the potential gains in MAE fluctuate between the different wind farms. These fluctuations could be attributed to the fundamental difference between analytical wake models and machine learning methods. While the former explicitly models the wake effects within the farm, the latter implicitly accounts for all patterns in the data affecting the power output. Therefore, machine learning methods can also consider external factors such as active curtailment strategies, which may be present in the operational context of certain wind farms. When estimating the potential power generation rather than the actual production, analytical wake models could provide an alternative to filtering out observations with curtailments from the dataset. 

\begin{table*}[t!]
\caption{Normalized continuous rank probability score per wind farm, as well as the average score over all wind farms. The values between brackets indicate the normalized CRPS on the training data. The metric is normalized by the installed capacity of the wind farm. Lower scores indicate a better fit, with the best scores highlighted in bold.}
\label{table:crps}
\begin{center}
\resizebox{0.75\textwidth}{!}{
    \begin{tabular}{lccccc}
        \toprule
         & SVGP & NGBoost & CQR & \multicolumn{2}{c}{Treeffuser} \\
         & & (Gaussian) & & (no tuning) & (tuned) \\
        \midrule
        Belwind       & 5.6\% (5.4\%) & 5.2\% (4.6\%) & 5.0\% (4.7\%) & 5.2\% (1.6\%) & \textbf{4.8\%} (4.1\%) \\
        C-Power       & 5.8 \% (5.5\%) & 5.4\% (4.6\%) & 5.0\% (4.1\%) & 5.2\% (1.6\%) & \textbf{4.9\%} (3.5\%) \\
        Mermaid       & 6.6\% (6.3\%) & 6.5\% (5.4\%) & 6.3\% (5.3\%) & 6.8\% (1.9\%) & \textbf{6.2\%} (4.8\%) \\
        Nobelwind     & 6.4\% (6.2\%) & 6.0\% (5.3\%) & 5.7\% (5.1\%) & 6.0\% (1.9\%) & \textbf{5.6\%} (4.6\%) \\
        Norther       & 6.5\% (6.6\%) & 6.1\% (5.1\%) & 5.8\% (5.0\%) & 5.9\% (1.8\%) & \textbf{5.6\%} (3.6\%) \\
        Northwester 2 & \textbf{6.1}\% (5.9\%) & 6.5\% (5.1\%) & \textbf{6.1\%} (5.0\%) & 6.6\% (2.0\%) & \textbf{6.1\%} (5.2\%) \\
        Northwind     & 6.3\% (6.0\%) & 5.7\% (5.2\%) & 5.4\% (4.7\%) & 5.6\% (1.7\%) & \textbf{5.2\%} (4.4\%) \\
        Rentel        & 6.6\% (5.9\%) & 6.3\% (4.8\%) & 6.1\% (4.6\%) & 6.5\% (1.7\%) & \textbf{6.0\%} (4.4\%) \\
        Seastar       & 6.4\% (6.3\%) & 6.1\% (5.3\%) & \textbf{5.8\%} (5.1\%) & 6.2\% (2.0\%) & \textbf{5.8\%} (5.2\%) \\
        \midrule
        Average       & 6.3\% (6.0\%) & 6.0\% (5.0\%) & 5.7\% (4.8\%) & 6.0\% (1.8\%) & \textbf{5.6\%} (4.4\%) \\
        \bottomrule
    \end{tabular}
}
\end{center}
\end{table*}
In Table \ref{table:crps}, the normalized continuous rank probability scores on the train and test set are shown for all wind farms.  In accordance with the results presented in in Table \ref{table:mae}, the Treeffuser model with optimized hyperparameters achieves the best out-of-sample performance. Conversely, we find that all tree-based models outperform Gaussian process regression in terms of probabilistic skill. In addition, the Treeffuser model without hyperparameter tuning now ranks among the worst performers, which can be attributed to overfitting on the training data. This conflicts with the statement in \cite{beltran2024treeffuser} that hyperparameter tuning was not required for this method. For this reason, we do not consider this model in the remainder of the paper. Additionally, the natural gradient boosting model is again among the worst performers, which could indicate that the Normal distribution is not a good fit to model the conditional power distribution. This hypothesis is further supported by the fact that SVGP also assumes a Normal distribution and achieves similar performance.

\begin{table}[t!]
\caption{Evaluation metrics per operating region of the power curve averaged over the different wind farms. Normalization is done using the installed capacity of the farm. Lower scores indicate a better fit, with the best scores highlighted in bold.}
\label{table:score_per_region}
\centering
\subcaption{Normalized mean absolute error}
\begin{tabular}{lccccc}
    \toprule
     & Wake & SVGP & NGBoost & CQR & Treeffuser \\
     & model & & (Gaussian) & & \\
    \midrule
    Region 1 & 0.8\% & 1.7\% & 1.7\% & 0.8\% & \textbf{0.6\%} \\
    Region 2 & 13.3\% & 9.0\% & 9.2\% & 9.0\% & \textbf{8.9\%} \\
    Region 3 & 11.5\% & 8.5\% & 8.8\% & 8.6\% & \textbf{8.2\%} \\
    \bottomrule
\end{tabular}
\bigskip
\subcaption{Normalized continuous ranked probability score}
\begin{tabular}{lccccc}
    \toprule
     & Wake & SVGP & NGBoost & CQR & Treeffuser \\
     & model & & (Gaussian) & & \\
    \midrule
    Region 1 & NA & 3.0\% & 1.4\% & 0.7\% & \textbf{0.6\%} \\
    Region 2 & NA & 6.6\% & 6.4\% & 6.2\% & \textbf{6.1\%} \\
    Region 3 & NA & 6.2\% & 6.5\% & 6.1\% & \textbf{5.9\%} \\
    \bottomrule
\end{tabular}
\end{table}
\begin{figure*}[hp!]
    \centering
    \begin{subfigure}{0.3\textwidth}
        \centering
        \includegraphics[width=\linewidth]{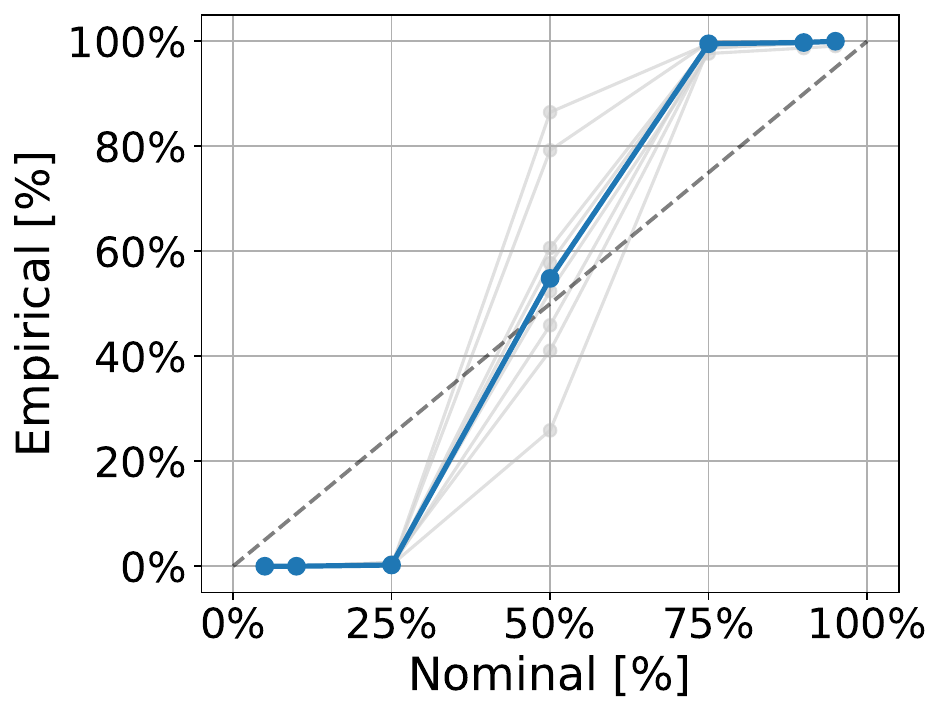}
        \caption{SVGP: Region 1}
    \end{subfigure}
    \hfill
    \begin{subfigure}{0.3\textwidth}
        \centering
        \includegraphics[width=\linewidth]{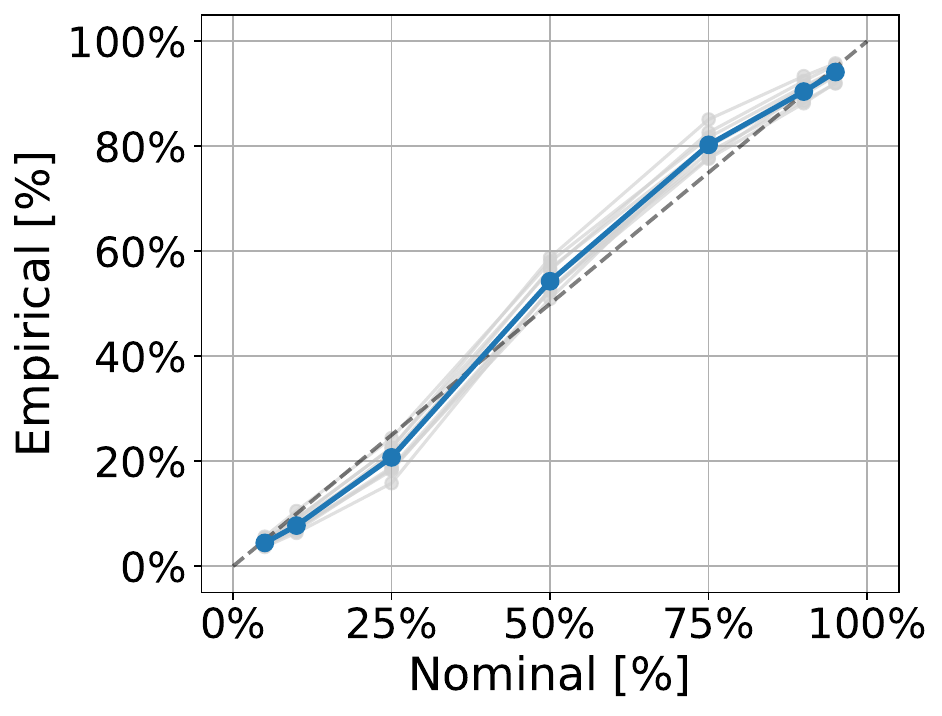}
        \caption{SVGP: Region 2}
    \end{subfigure}
    \hfill
    \begin{subfigure}{0.3\textwidth}
        \centering
        \includegraphics[width=\linewidth]{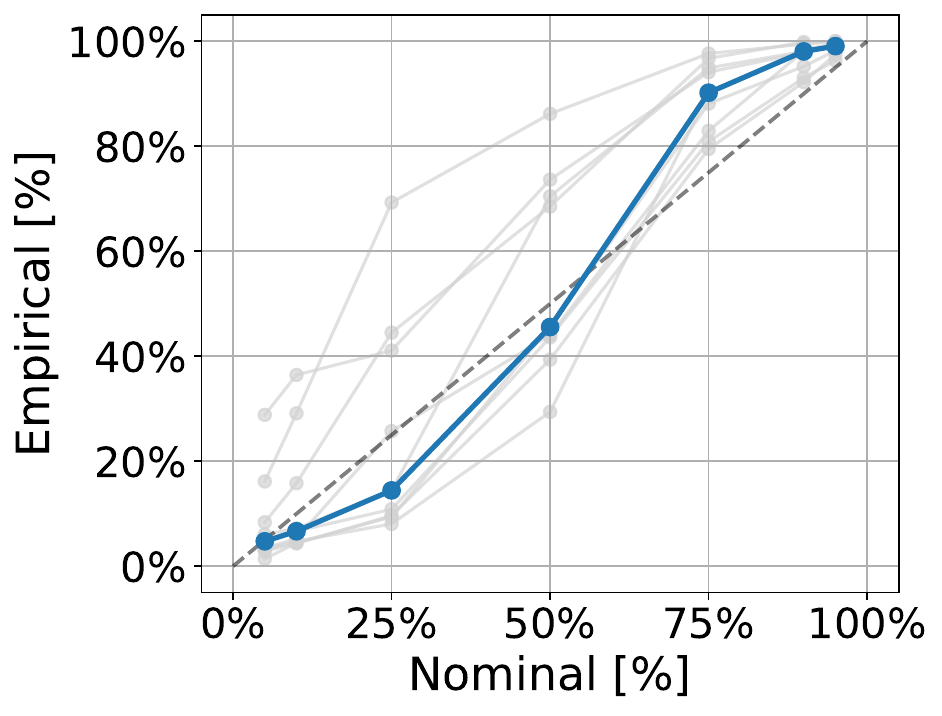}
        \caption{SVGP: Region 3}
    \end{subfigure}
    \vspace{12pt}
    \begin{subfigure}{0.3\textwidth}
        \centering
        \includegraphics[width=\linewidth]{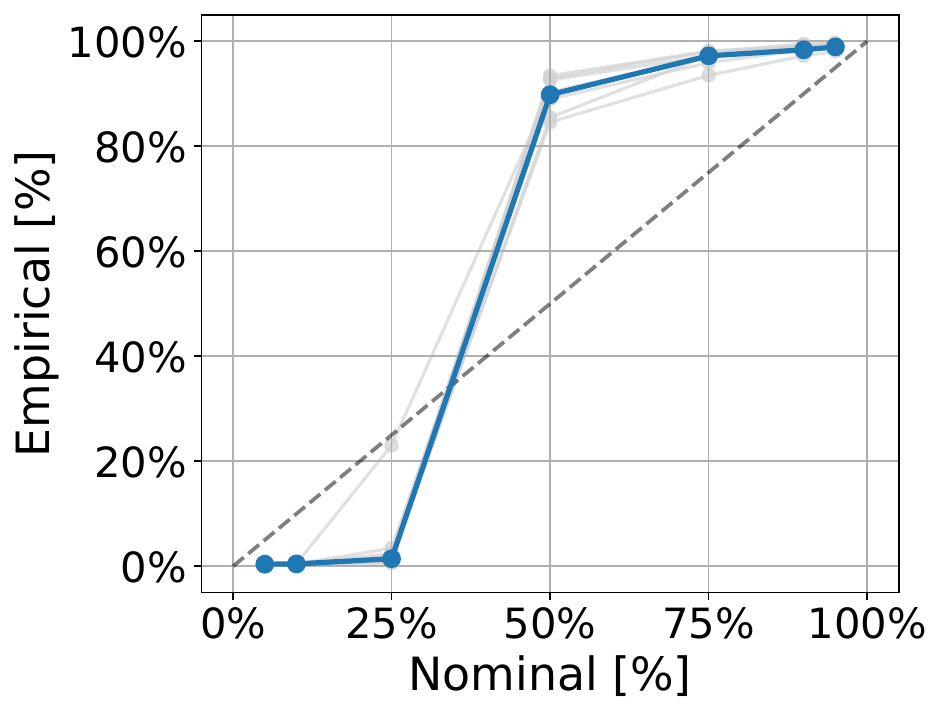}
        \caption{NGBoost: Region 1}
    \end{subfigure}
    \hfill
    \begin{subfigure}{0.3\textwidth}
        \centering
        \includegraphics[width=\linewidth]{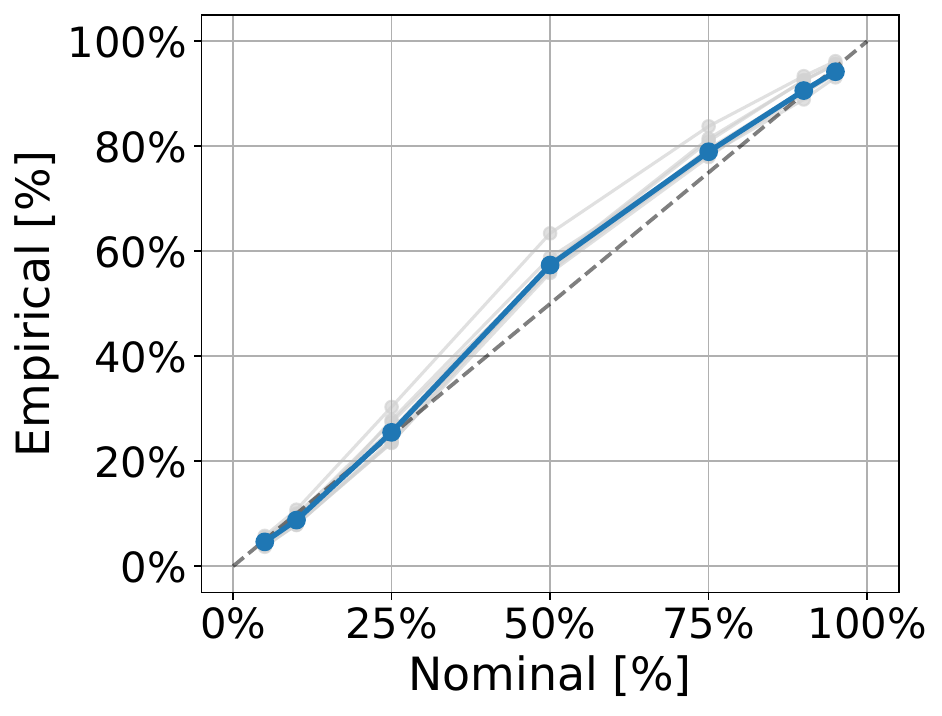}
        \caption{NGBoost: Region 2}
    \end{subfigure}
    \hfill
    \begin{subfigure}{0.3\textwidth}
        \centering
        \includegraphics[width=\linewidth]{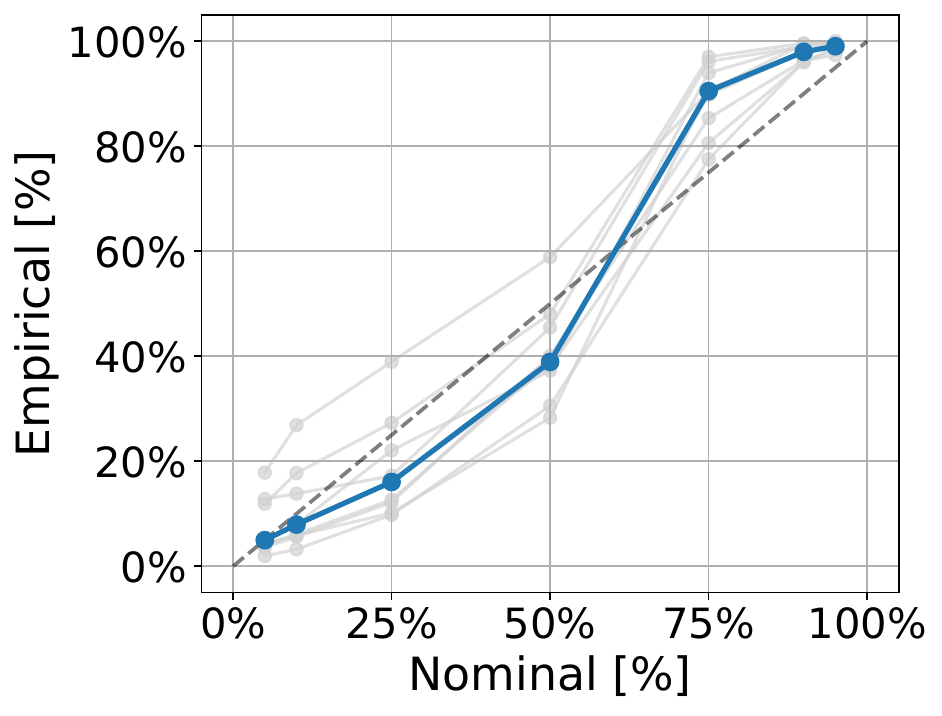}
        \caption{NGBoost: Region 3}
    \end{subfigure}
    \vspace{12pt}
    \begin{subfigure}{0.3\textwidth}
        \centering
        \includegraphics[width=\linewidth]{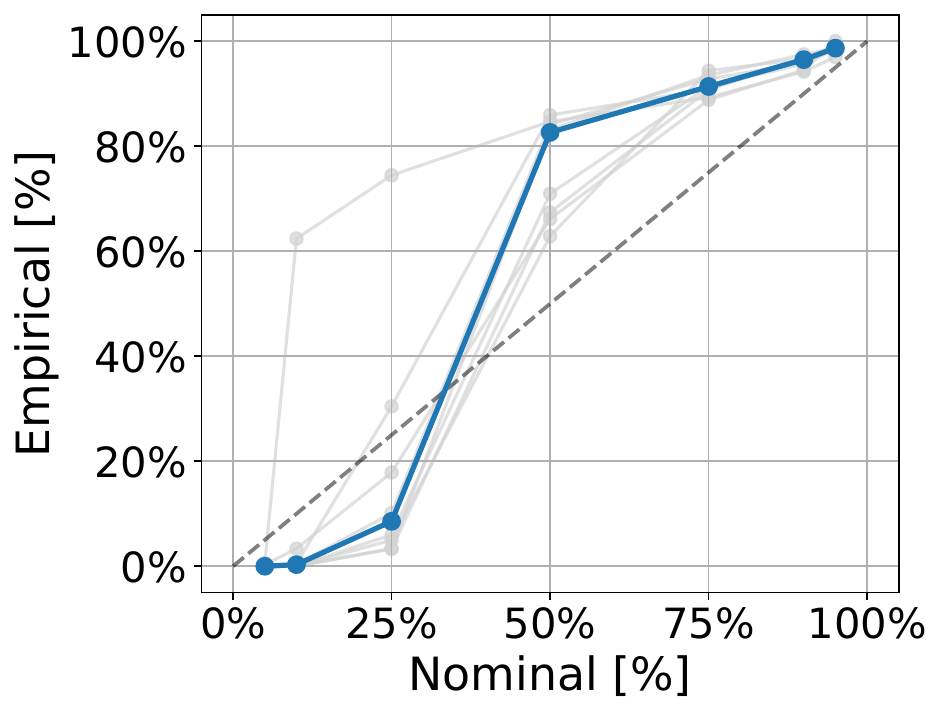}
        \caption{CQR: Region 1}
    \end{subfigure}
    \hfill
    \begin{subfigure}{0.3\textwidth}
        \centering
        \includegraphics[width=\linewidth]{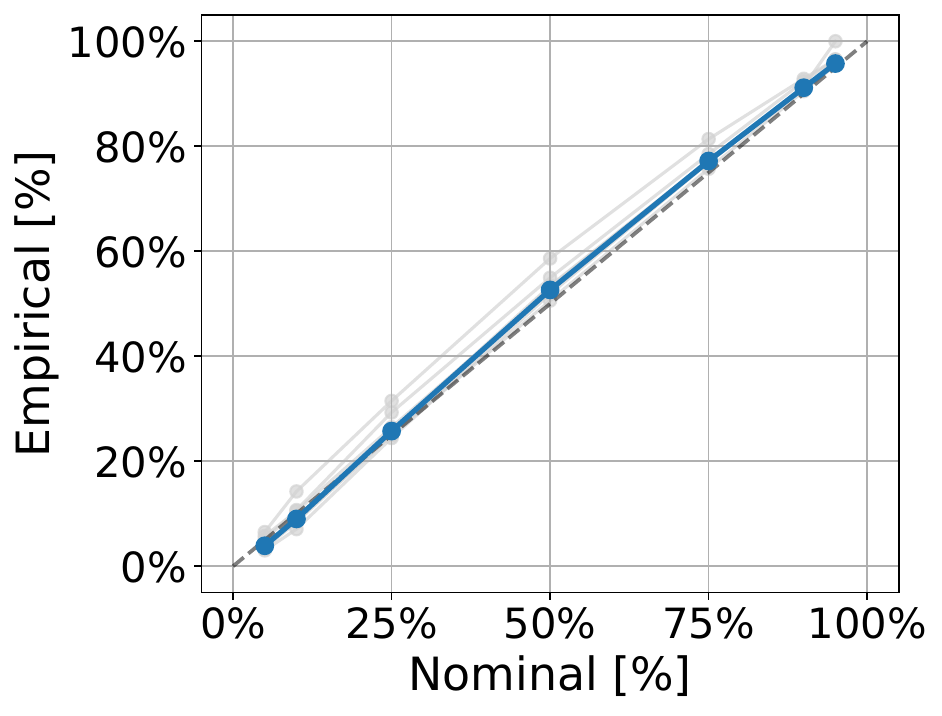}
        \caption{CQR: Region 2}
    \end{subfigure}
    \hfill
    \begin{subfigure}{0.3\textwidth}
        \centering
        \includegraphics[width=\linewidth]{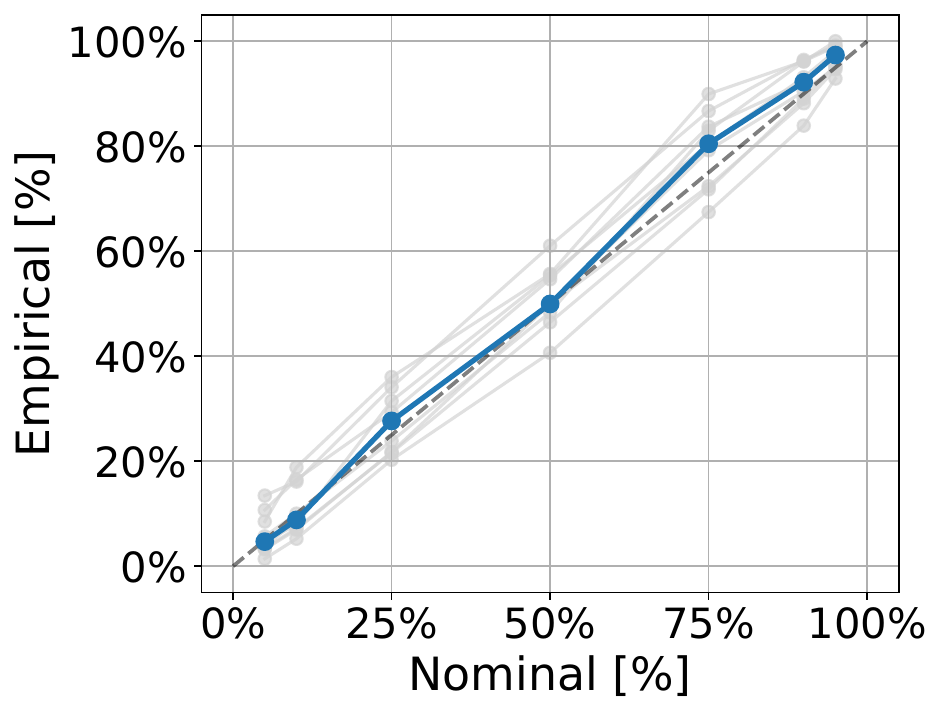}
        \caption{CQR: Region 3}
    \end{subfigure}
    \vspace{12pt}
    \begin{subfigure}{0.3\textwidth}
        \centering
        \includegraphics[width=\linewidth]{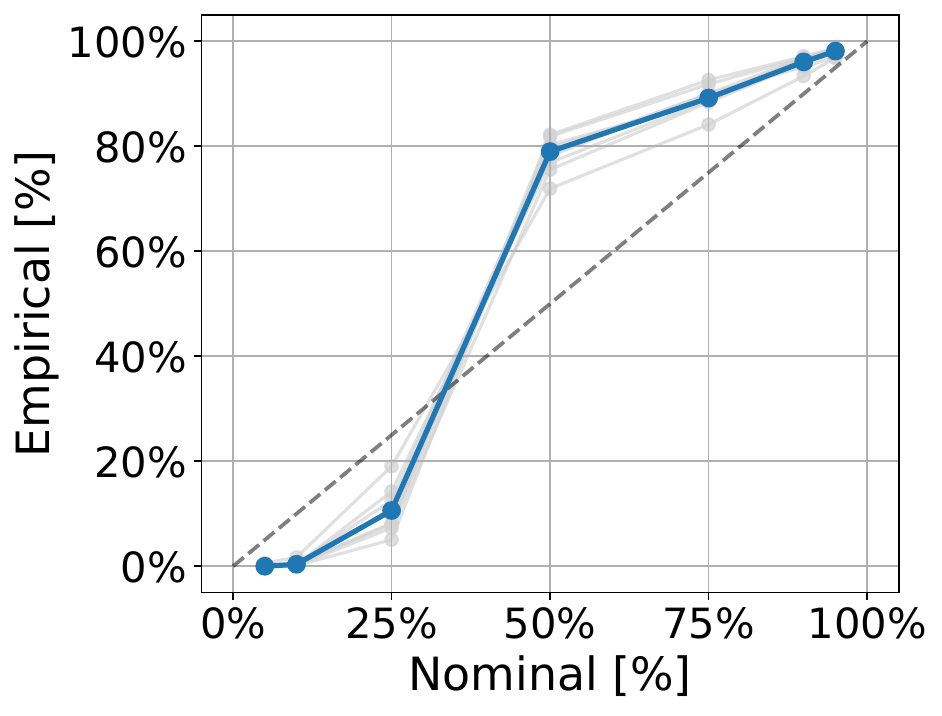}
        \caption{Treeffuser: Region 1}
    \end{subfigure}
    \hfill
    \begin{subfigure}{0.3\textwidth}
        \centering
        \includegraphics[width=\linewidth]{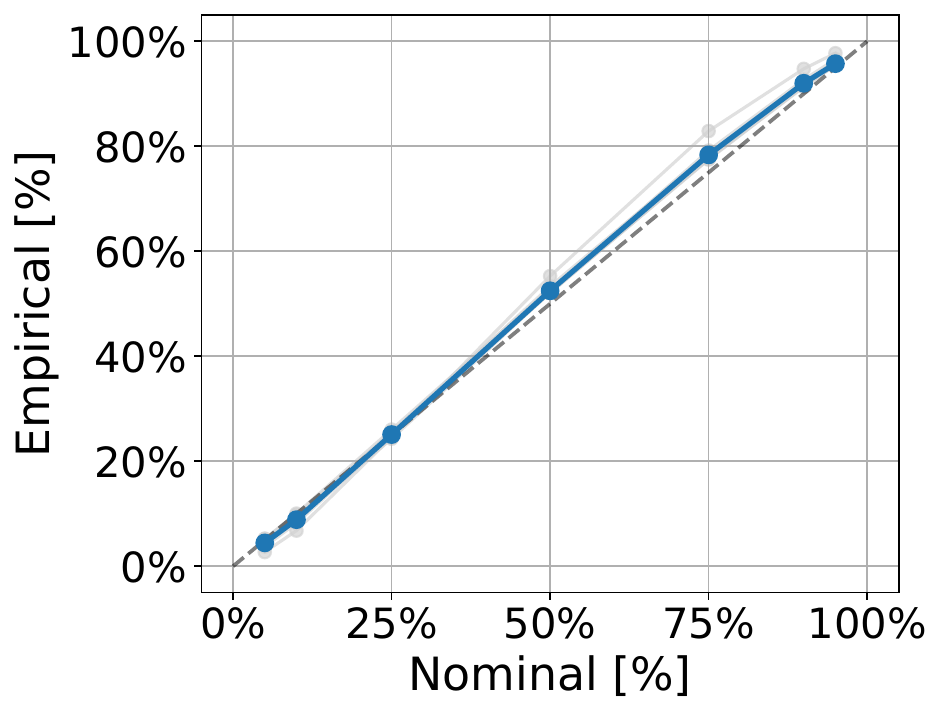}
        \caption{Treeffuser: Region 2}
    \end{subfigure}
    \hfill
    \begin{subfigure}{0.3\textwidth}
        \centering
        \includegraphics[width=\linewidth]{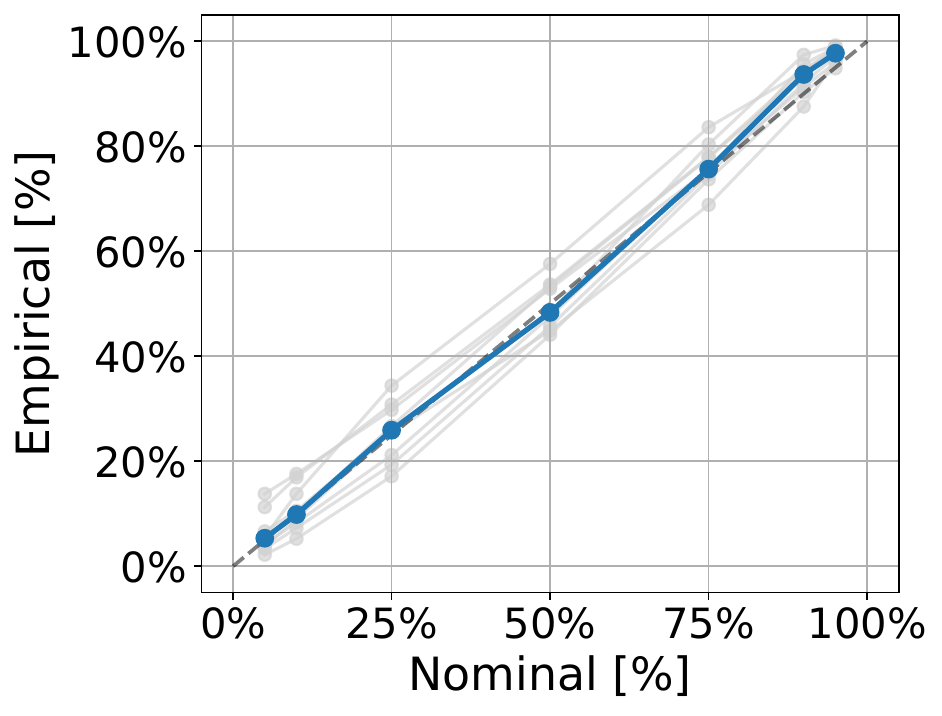}
        \caption{Treeffuser: Region 3}
    \end{subfigure}
    \caption{Reliability curves for the probabilistic forecasting models per operating region of the power curve. The blue line shows the median for all wind farms, while the grey lines indicates the reliability for each individual farm. The dashed grey line represents ideal calibration, curves closer to this line are better calibrated. When the curves are positioned above the diagonal, this indicates that the predictions are conservative (i.e., over-coverage), while those positioned below the diagonal signal under-coverage.}
    \label{fig:reliability_curves_per_region}
\end{figure*}
\begin{figure}[ht!]
    \centering
    \begin{subfigure}{\linewidth}
        \centering
        \includegraphics[width=0.7\linewidth]{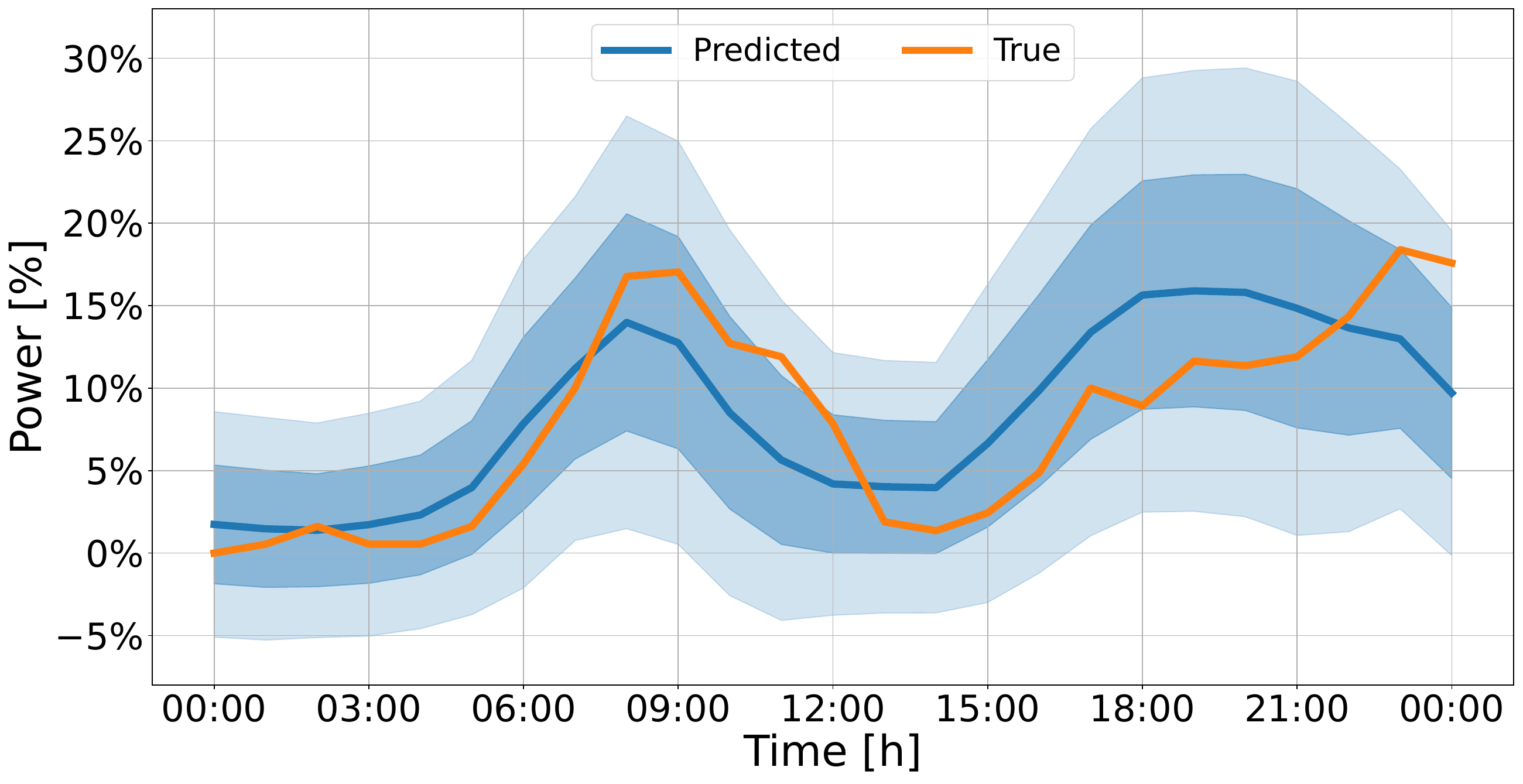}
        \caption{Natural gradient boosting}
        \vspace{4pt}
    \end{subfigure}
    \begin{subfigure}{\linewidth}
        \centering
        \includegraphics[width=0.7\linewidth]{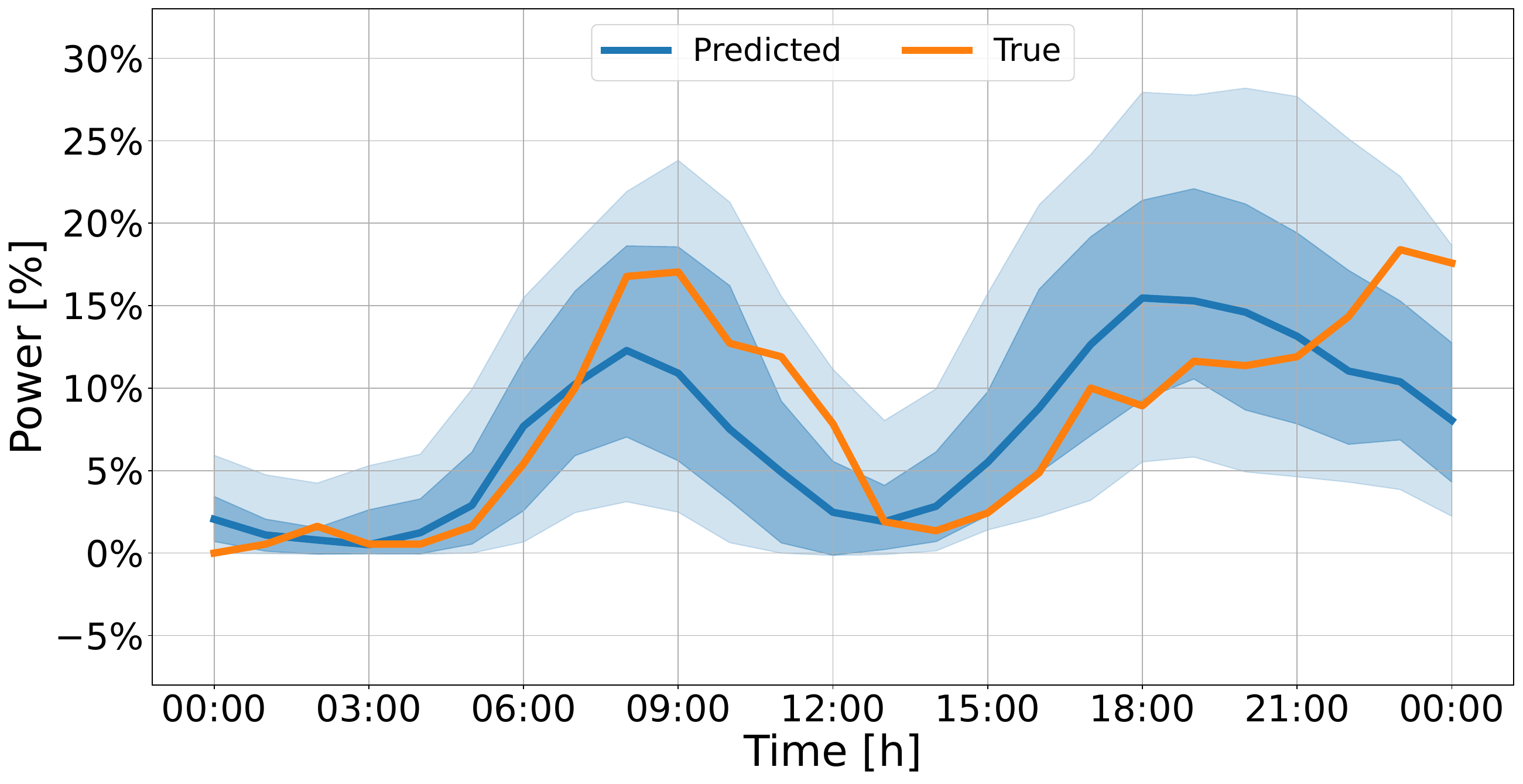}
        \caption{Conformalized quantile regression}
        \vspace{4pt}
    \end{subfigure}
    \begin{subfigure}{\linewidth}
        \centering
        \includegraphics[width=0.7\linewidth]{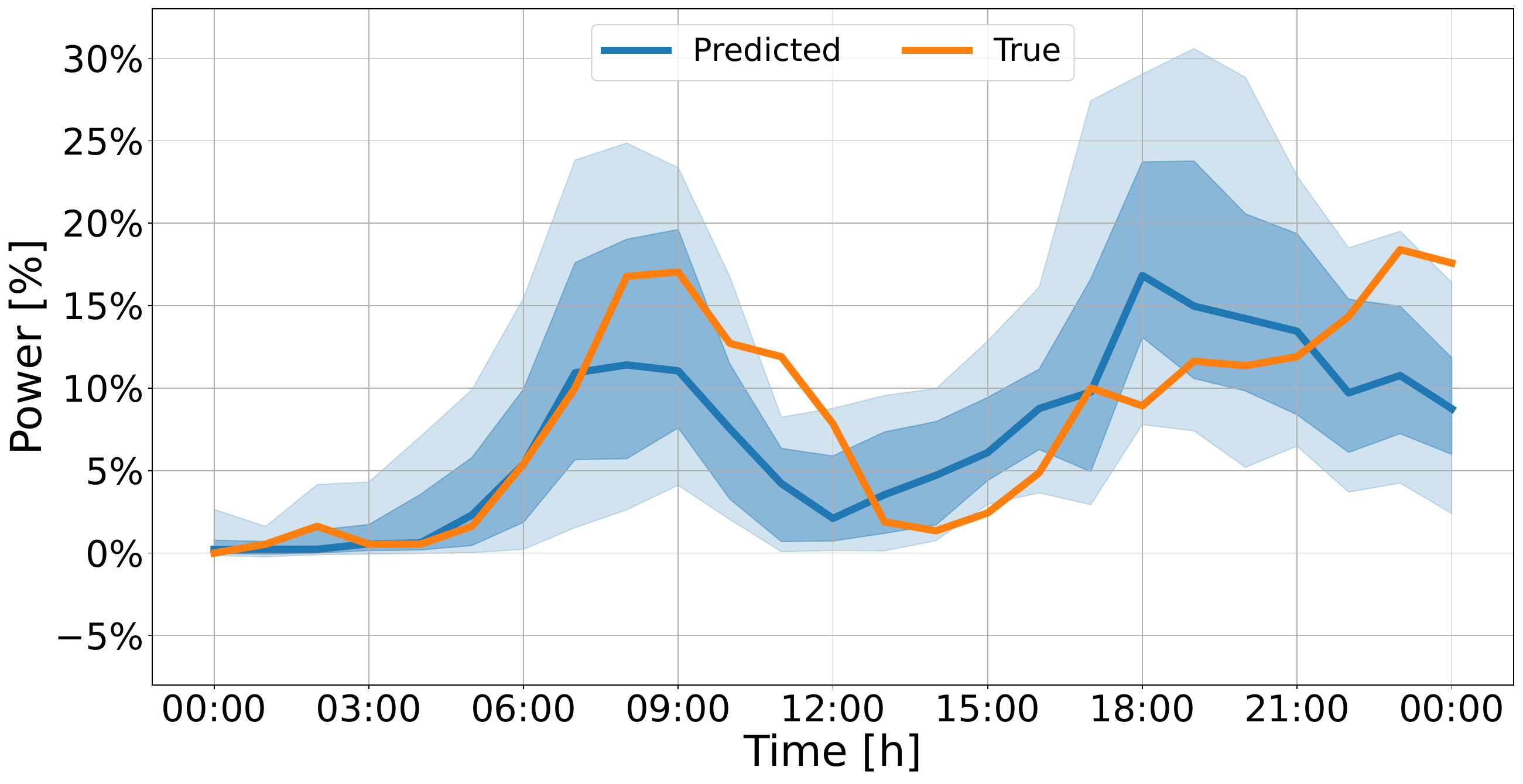}
        \caption{Conditional diffusion model (Treeffuser)}
    \end{subfigure}
    \caption{Illustrative example of the day-ahead forecasts using the different tree-based methods, when the wind speed forecasts are around the cut-in wind speed throughout the day. The dark and the light blue area represent, respectively, the 50\% and the 80\% confidence intervals. The natural gradient boosting model yields confidence intervals with negative power values because of the symmetry of the assumed normal distribution. This highlights a limitation of parametric methods for wind power forecasting, since the output is physically constrained to be non-negative. This issue is not observed in the other models, which are non-parametric and therefore more flexible to model the true data distribution.}
    \label{fig:example}
\end{figure}
Table \ref{table:score_per_region} shows the average score across the wind farms per operating region of the power curve. The operating region is determined based on the average wind speed forecast. Traditionally, four different operating regions are defined based on the warranted power curve \cite{saint2020parametric}: The first region covers wind speeds below the cut-in speed. The second region includes all wind speeds between the cut-in and rated wind speeds. The third region is defined by wind speeds ranging from the rated wind speed to the cut-out wind speed. The final region encompasses all wind speeds above the cut-out wind speed. However, region four is excluded from the analysis due to a lack of sufficient observations. This region is also strongly associated with storm events and cut-out events, which are known to cause the largest errors in day-ahead wind power forecasts \cite{steiner2017critical}. Consequently, specialized storm forecasting tools, such as \cite{smet2023probabilistic}, are often employed for these conditions. The largest discrepancy between the two models assuming a Gaussian distribution and the remaining models is observed in operating regions 1 and 3. This supports our belief that the Normal distribution is not a good fit to model the conditional power distribution, since in these operating regions the power output is not symmetrically distributed (see, for example, \cite{lange2005uncertainty, kou2013sparse}). The two other methods are distribution-free, meaning that they do not assume an underlying parametric distribution, and therefore they can provide a probability distribution of any shape. Figure \ref{fig:reliability_curves_per_region} shows the reliability curves per operating region, evaluating empirical quantile coverage against the corresponding theoretical levels. We observe that, across all regions, the models assuming a Normal distribution are less well-calibrated compared to the others. For regions 1 and 3, the difference is again more pronounced. Additionally, the results show that, for all models, operating regions 1 and 3 are more challenging to calibrate, with operating region 1 being the most difficult. This can be attributed to the non-linear shape of the power curve in these regions. To further emphasize the differences between the tree-based models, Figure \ref{fig:example} provides a visual example of the forecasts for a day on which the average wind speed forecast was around the cut-in wind speed. The confidence interval of the natural gradient boosting model contains negative power outputs, which is physically not feasible, whereas for the other two methods this issue does not occur.

\begin{figure}[ht!]
    \centering
    \includegraphics[width=0.9\linewidth]{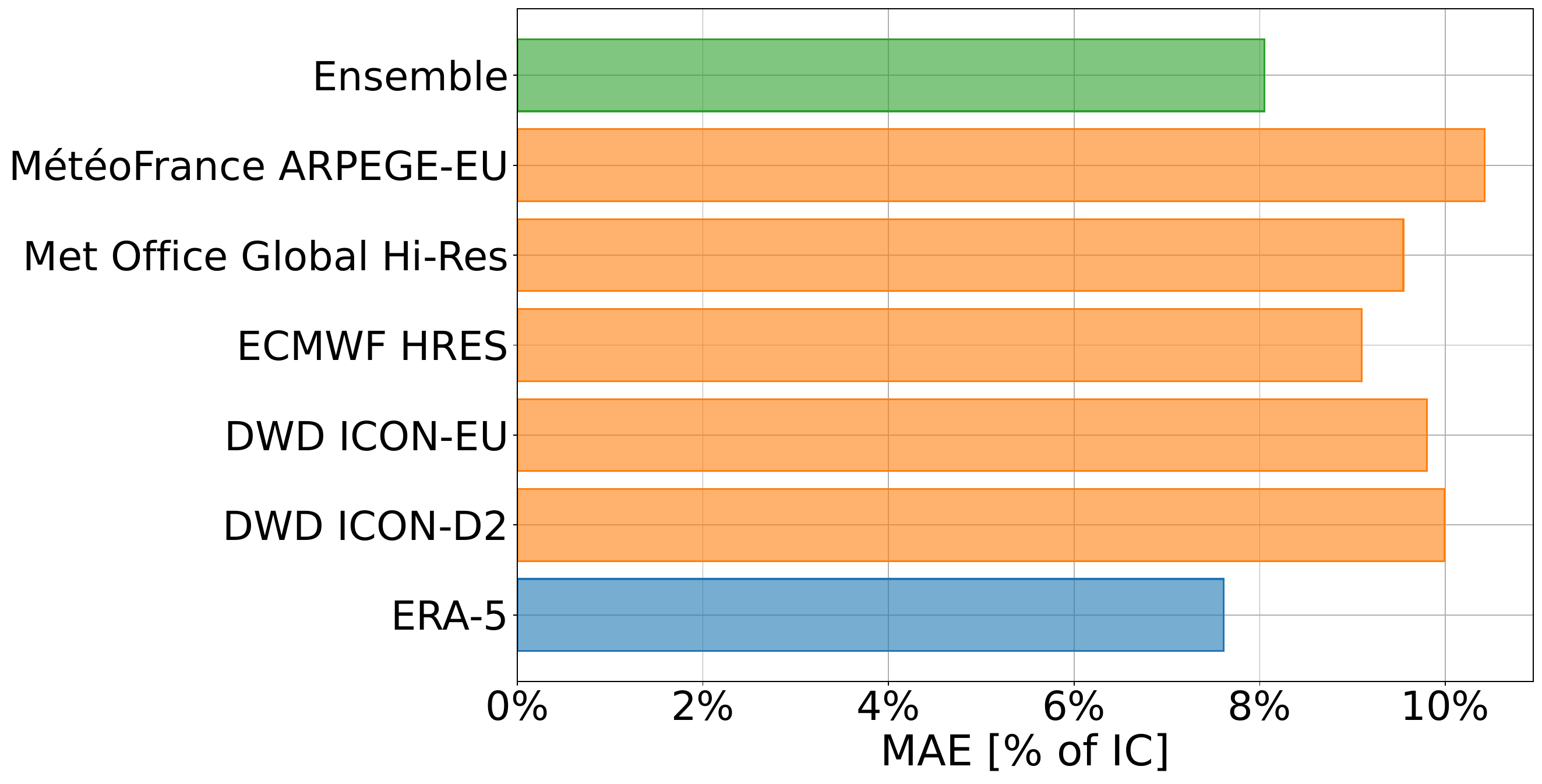}
    \caption{Average mean absolute error (as a \% of the installed capacity of the wind farm) for the Treeffuser model over the different wind farms with different input datasets: ensemble of weather forecasts (green), single weather forecast (orange), re-analysis weather data (blue).}
    \label{fig:mae_wf}
\end{figure}

In Figure \ref{fig:mae_wf}, we illustrate how the use of an ensemble of weather forecasts impacts the MAE of the forecasting models. To this end, we trained the Treeffuser model with different input data: the ensemble of weather forecasts, the individual forecasts and ERA-5 reanalysis data. The reanalysis data is used to assess the impact of the weather forecast error, as it is derived using data assimilation by combining actual measurements with a numerical weather model. For each input dataset, the model was trained for each wind farm and the average MAE was taken over all wind farms. The results demonstrate that the ensemble model substantially improves predictive performance relative to the individual forecasts, even attaining an MAE close to the value obtained using the reanalysis wind speed and direction from ERA-5. Compared to the average single-provider forecast, the ensemble method achieves an improvement of $17\%$ in MAE, and up to $23\%$ compared to the worst-performing provider. In \ref{app:ensemble_analysis}, we also provide a per-farm breakdown of the error as well as Diebold-Mariano tests \cite{diebold2002comparing} to prove the statistical significance of the results.

Table \ref{tab:model_efficiency} shows the average training and inference times for the different models. The natural gradient boosting model achieves the fastest inference time, as it requires only two base learners, one per parameter of the Normal distribution, compared to one learner per predicted quantile for conformalized quantile regression. The conditional diffusion model is substantially slower during both training and inference. This is because training requires generating multiple noisy versions of the data, whereas inference relies on producing multiple samples per observation by numerically solving a stochastic differential equation \cite{beltran2024treeffuser}.

 \begin{table}[t!]
\centering
\caption{Average computation time for the different models across all wind farms. Training time is reported using the optimal hyperparameter configuration. Inference time is reported per observation. The number of CPU cores used by each method is also shown.}
\label{tab:model_efficiency}
    \begin{tabular}{lccc}
        \toprule
         & NGBoost (Gaussian) & CQR & Treeffuser \\
        \midrule
        Training time  & $15.6s$ & $49.6s$ & $119s$ \\
        Inference time & $9.5\mu s$ & $61.2\mu s$ & $0.3s$ \\
        CPU cores      & $1$     & $1$    & $2$   \\
        \bottomrule
    \end{tabular}
\end{table}

In summary, the empirical results indicate that tree-based machine learning models significantly outperform engineering methods for day-ahead forecasting of wind power. These methods can achieve an improved point forecast accuracy of up to 33\% and 53\% compared to the power curve and the analytical wake model, while also providing an estimate of the predictive distribution. Furthermore, all tree-based models outperform Gaussian process regression in terms of continuous rank probability score, while the distribution-free methods also improve point forecast accuracy. The underperformance of the natural gradient boosting model might be attributed to the Gaussian likelihood assumption, which is not suited for operating regions where the power output distribution is asymmetric. An interesting pursuit for future research could be to extend natural gradient boosting to other parametric distributions, such as a mixture of univariate Gaussian densities or a Beta distribution. Moreover, the findings suggest that incorporating an ensemble of weather forecasts significantly enhances predictive accuracy relative to relying on a single forecast provider. This represents another advantage compared to engineering methods, which presume the wind speed and direction to be known and cannot handle multiple weather inputs. Nonetheless, the analytical wake model already shows a considerable improvement compared to the power curve and can be a good alternative to machine learning methods when interpretability is more important and probabilistic estimates are not required. Furthermore, the analytical wake models can provide an unbiased estimation of potential power generation when the wind farm pursues an active curtailment strategy. Among the different probabilistic prediction methods, we found that conditional diffusion model attains the best overall point and probabilistic estimate of wind power generation across all wind farms. On the other hand, we observed a significantly higher computation time for training and inference when using this method in our experiments. Conformalized quantile regression obtained good performance at a lower computation time and can therefore provide a valuable alternative to conditional diffusion when computation time is important. However, day-ahead forecasting only requires a limited number of predictions per day and therefore the most accurate model remains the most suitable in this context. 

\section{Conclusion}
\label{sec:conclusion}

This work studied three different methods for probabilistic day-ahead forecasting of wind power generation using tree-based machine learning and an ensemble of weather forecasts. To this end, we conducted a comparative analysis with data from all Belgian wind farms over a period of four years. Alongside the tree-based models, we also considered deterministic engineering methods, using either the power curve or an analytical wake model, as well as a probabilistic baseline using stochastic variational Gaussian process regression. 

Our findings showed that the all models significantly reduce point forecast error compared to the deterministic engineering baselines. One possible explanation is the ability of data-driven methods to incorporate an ensemble of weather forecasts, which was found to considerably improve predictive performance. Nonetheless, the results also show that analytical wake models present a good alternative to machine learning methods when interpretability is more important and probabilistic forecasts are not required. Compared to the probabilistic baseline, all three methods outperformed in terms of probabilistic skill, while the distribution-free methods also improved point forecast accuracy. Among the tree-based models, the conditional diffusion model yielded the most accurate probabilistic and point estimates of wind power generation.

\section*{Acknowledgments}

The authors gratefully acknowledge the financial support of the Flemish Government through the Flanders AI Research Program, the Energy Transition Funds of the Belgian Federal Government through the BeFORECAST project and the Sustainable Blue Economy Partnership through the INSPIRE project (project no. SBEP2023-440).

\FloatBarrier
\printbibliography

@article{nowotarski2018recent,
  title={Recent advances in electricity price forecasting: A review of probabilistic forecasting},
  author={Nowotarski, Jakub and Weron, Rafa{\l}},
  journal={Renewable and Sustainable Energy Reviews},
  volume={81},
  pages={1548--1568},
  year={2018},
  publisher={Elsevier}
}

@article{matheson1976scoring,
  title={Scoring rules for continuous probability distributions},
  author={Matheson, James E and Winkler, Robert L},
  journal={Management science},
  volume={22},
  number={10},
  pages={1087--1096},
  year={1976},
  publisher={INFORMS}
}

@article{hersbach2000decomposition,
  title={Decomposition of the continuous ranked probability score for ensemble prediction systems},
  author={Hersbach, Hans},
  journal={Weather and Forecasting},
  volume={15},
  number={5},
  pages={559--570},
  year={2000}
}

@online{entsoe,
  title   = {{ENTSO-E} Transparency Platform},
  author  = {{European Network of Transmission System Operators for Electricity}},
  url     = {https://transparency.entsoe.eu/},
  urldate = {2025-07-01},
}

@article{hou2021data,
  title={Data-driven robust day-ahead unit commitment model for hydro/thermal/wind/photovoltaic/nuclear power systems},
  author={Hou, Wenting and Wei, Hua},
  journal={International Journal of Electrical Power \& Energy Systems},
  volume={125},
  pages={106427},
  year={2021},
  publisher={Elsevier}
}

@article{hou2022hybrid,
  title={A hybrid data-driven robust optimization approach for unit commitment considering volatile wind power},
  author={Hou, Wenting and Hou, Linfang and Zhao, Shaohua and Liu, Wei},
  journal={Electric Power Systems Research},
  volume={205},
  pages={107758},
  year={2022},
  publisher={Elsevier}
}

@article{he2026short,
  title={Short-term prediction of wind vector at multi-heights via deep learning techniques based on marine measurements from Light-Detection-and-Ranging device},
  author={He, YC and Pu, JL and Gan, SL and Gan, ZQ and Yang, HW and Huang, YJ and Chan, PW and Fu, JY},
  journal={Physics of Fluids},
  volume={38},
  number={1},
  year={2026},
  publisher={AIP Publishing}
}

@inproceedings{10.5555/3023638.3023667,
author = {Hensman, James and Fusi, Nicol\`{o} and Lawrence, Neil D.},
title = {Gaussian processes for Big data},
year = {2013},
publisher = {AUAI Press},
address = {Arlington, Virginia, USA},
booktitle = {Proceedings of the Twenty-Ninth Conference on Uncertainty in Artificial Intelligence},
pages = {282–290},
numpages = {9},
location = {Bellevue, WA},
series = {UAI'13}
}

@article{diebold2002comparing,
  title={Comparing predictive accuracy},
  author={Diebold, Francis X and Mariano, Robert S},
  journal={Journal of Business \& economic statistics},
  volume={20},
  number={1},
  pages={134--144},
  year={2002},
  publisher={Taylor \& Francis}
}

@inproceedings{hensman2015scalable,
  title={Scalable variational Gaussian process classification},
  author={Hensman, James and Matthews, Alexander and Ghahramani, Zoubin},
  booktitle={Proceedings of the Eighteenth International Conference on Artificial Intelligence and Statistics},
  pages={351--360},
  year={2015},
  organization={PMLR}
}

@article{kou2013sparse,
  title={Sparse online warped Gaussian process for wind power probabilistic forecasting},
  author={Kou, Peng and Gao, Feng and Guan, Xiaohong},
  journal={Applied energy},
  volume={108},
  pages={410--428},
  year={2013},
  publisher={Elsevier}
}

@misc{Zippenfenig_Open-Meteo,
  author = {Zippenfenig, Patrick},
  doi = {10.5281/zenodo.7970649},
  licence = {CC-BY-4.0},
  title = {Open-Meteo.com Weather API},
  year = {2023},
  copyright = {Creative Commons Attribution 4.0 International},
  url = {https://open-meteo.com/}
}

@inproceedings{van2024performance,
  title={Performance comparison of analytical wake models calibrated on a large offshore wind cluster},
  author={Van Binsbergen, Diederik and Daems, Pieter-Jan and Verstraeten, Timothy and Nejad, Amir and Helsen, Jan},
  booktitle={Journal of Physics: Conference Series},
  volume={2767},
  number={9},
  pages={092059},
  year={2024},
  organization={IOP Publishing}
}

@article{gonzalez2012wake,
  title={Wake effect in wind farm performance: Steady-state and dynamic behavior},
  author={Gonz{\'a}lez-Longatt, Francisco and Wall, Peter and Terzija, Vladimir},
  journal={Renewable Energy},
  volume={39},
  number={1},
  pages={329--338},
  year={2012},
  publisher={Elsevier}
}

@article{beltran2024treeffuser,
  title={Treeffuser: probabilistic prediction via conditional diffusions with gradient-boosted trees},
  author={Beltran Velez, Nicolas and Grande, Alessandro A and Nazaret, Achille and Kucukelbir, Alp and Blei, David},
  journal={Advances in Neural Information Processing Systems},
  volume={37},
  pages={118296--118325},
  year={2024}
}

@article{romano2019conformalized,
  title={Conformalized quantile regression},
  author={Romano, Yaniv and Patterson, Evan and Candes, Emmanuel},
  journal={Advances in neural information processing systems},
  volume={32},
  year={2019}
}

@article{chernozhukov2010quantile,
  title={Quantile and probability curves without crossing},
  author={Chernozhukov, Victor and Fern{\'a}ndez-Val, Iv{\'a}n and Galichon, Alfred},
  journal={Econometrica},
  volume={78},
  number={3},
  pages={1093--1125},
  year={2010},
  publisher={Wiley Online Library}
}

@article{koenker1978regression,
  title={Regression quantiles},
  author={Koenker, Roger and Bassett Jr, Gilbert},
  journal={Econometrica: journal of the Econometric Society},
  pages={33--50},
  year={1978},
  publisher={JSTOR}
}

@article{angelopoulos2021gentle,
  title={A gentle introduction to conformal prediction and distribution-free uncertainty quantification},
  author={Angelopoulos, Anastasios N and Bates, Stephen},
  journal={arXiv preprint arXiv:2107.07511},
  year={2021}
}

@inproceedings{duan2020ngboost,
  title={Ngboost: Natural gradient boosting for probabilistic prediction},
  author={Duan, Tony and Anand, Avati and Ding, Daisy Yi and Thai, Khanh K and Basu, Sanjay and Ng, Andrew and Schuler, Alejandro},
  booktitle={International conference on machine learning},
  pages={2690--2700},
  year={2020},
  organization={PMLR}
}

@inproceedings{song2021scorebased,
  author       = {Yang Song and
                  Jascha Sohl{-}Dickstein and
                  Diederik P. Kingma and
                  Abhishek Kumar and
                  Stefano Ermon and
                  Ben Poole},
  title        = {Score-Based Generative Modeling through Stochastic Differential Equations},
  booktitle    = {9th International Conference on Learning Representations, {ICLR} 2021,
                  Virtual Event, Austria, May 3-7, 2021},
  publisher    = {OpenReview.net},
  year         = {2021},
  url          = {https://openreview.net/forum?id=PxTIG12RRHS},
  bibsource    = {dblp computer science bibliography, https://dblp.org}
}

@article{batzolis2021conditional,
  title={Conditional image generation with score-based diffusion models},
  author={Batzolis, Georgios and Stanczuk, Jan and Sch{\"o}nlieb, Carola-Bibiane and Etmann, Christian},
  journal={arXiv preprint arXiv:2111.13606},
  year={2021}
}

@article{saint2020parametric,
  title={A parametric model for wind turbine power curves incorporating environmental conditions},
  author={Saint-Drenan, Yves-Marie and Besseau, Romain and Jansen, Malte and Staffell, Iain and Troccoli, Alberto and Dubus, Laurent and Schmidt, Johannes and Gruber, Katharina and Sim{\~o}es, Sofia G and Heier, Siegfried},
  journal={Renewable Energy},
  volume={157},
  pages={754--768},
  year={2020},
  publisher={Elsevier}
}

@article{lange2005uncertainty,
  title={On the uncertainty of wind power predictions—Analysis of the forecast accuracy and statistical distribution of errors},
  author={Lange, Matthias},
  journal={J. Sol. Energy Eng.},
  volume={127},
  number={2},
  pages={177--184},
  year={2005}
}

@misc{IEA_GlobalEnergyReview2025,
  author = {IEA},
  title = {Global Energy Review 2025},
  year = {2025},
  publisher = {IEA},
  address = {Paris},
  url = {https://www.iea.org/reports/global-energy-review-2025},
  note = {Licence: CC BY 4.0}
}

@article{heptonstall2021systematic,
  title={A systematic review of the costs and impacts of integrating variable renewables into power grids},
  author={Heptonstall, Philip J and Gross, Robert JK},
  journal={Nature Energy},
  volume={6},
  number={1},
  pages={72--83},
  year={2021},
  publisher={Nature Publishing Group UK London}
}

@article{HONG2016896,
title = {Probabilistic energy forecasting: Global Energy Forecasting Competition 2014 and beyond},
journal = {International Journal of Forecasting},
volume = {32},
number = {3},
pages = {896-913},
year = {2016},
issn = {0169-2070},
doi = {https://doi.org/10.1016/j.ijforecast.2016.02.001},
url = {https://www.sciencedirect.com/science/article/pii/S0169207016000133},
author = {Tao Hong and Pierre Pinson and Shu Fan and Hamidreza Zareipour and Alberto Troccoli and Rob J. Hyndman},
}

@article{zhu2012short,
  title={Short-term wind speed forecasting for power system operations},
  author={Zhu, Xinxin and Genton, Marc G},
  journal={International Statistical Review},
  volume={80},
  number={1},
  pages={2--23},
  year={2012},
  publisher={Wiley Online Library}
}

@inproceedings{bruninx2025day,
  title={Day-Ahead Bidding Strategies for Wind Farm Operators Under a One-Price Balancing Scheme},
  author={Bruninx, Max and Verstraeten, Timothy and Kazempour, Jalal and Helsen, Jan},
  booktitle={Proceedings of the 16th ACM International Conference on Future and Sustainable Energy Systems},
  pages={719--726},
  year={2025}
}

@article{pinson2007trading,
  title={Trading wind generation from short-term probabilistic forecasts of wind power},
  author={Pinson, Pierre and Chevallier, Christophe and Kariniotakis, George N},
  journal={IEEE transactions on Power Systems},
  volume={22},
  number={3},
  pages={1148--1156},
  year={2007},
  publisher={IEEE}
}

@article{heiser2024betting,
  title={Betting vs. Trading: Learning a Linear Decision Policy for Selling Wind Power and Hydrogen},
  author={Heiser, Yannick and Pourahmadi, Farzaneh and Kazempour, Jalal},
  journal={Sustainable Energy, Grids and Networks},
  pages={101848},
  year={2025},
  publisher={Elsevier}
}

@article{pinson2013wind,
     author = {Pierre Pinson},
     journal = {Statistical Science},
     number = {4},
     pages = {564--585},
     publisher = {Institute of Mathematical Statistics},
     title = {Wind Energy: Forecasting Challenges for Its Operational Management},
     volume = {28},
     year = {2013}
}

@article{allen2002towards,
  title={Towards objective probabalistic climate forecasting},
  author={Allen, Myles R and Stainforth, David A},
  journal={Nature},
  volume={419},
  number={6903},
  pages={228--228},
  year={2002},
  publisher={Nature Publishing Group UK London}
}

@article{sweeney2020future,
  title={The future of forecasting for renewable energy},
  author={Sweeney, Conor and Bessa, Ricardo J and Browell, Jethro and Pinson, Pierre},
  journal={Wiley Interdisciplinary Reviews: Energy and Environment},
  volume={9},
  number={2},
  pages={e365},
  year={2020},
  publisher={Wiley Online Library}
}

@article{giebel2017wind,
  title={Wind power forecasting—a review of the state of the art},
  author={Giebel, Gregor and Kariniotakis, George},
  journal={Renewable energy forecasting},
  pages={59--109},
  year={2017},
  publisher={Elsevier}
}

@inproceedings{bellinguer2020probabilistic,
  title={Probabilistic forecasting of regional wind power generation for the eem20 competition: A physics-oriented machine learning approach},
  author={Bellinguer, Kevin and Mahler, Valentin and Camal, Simon and Kariniotakis, Georges},
  booktitle={2020 17th International Conference on the European Energy Market (EEM)},
  pages={1--6},
  year={2020},
  organization={IEEE}
}

@article{browell2025hybrid,
  title={The hybrid renewable energy forecasting and trading competition 2024},
  author={Browell, Jethro and van der Meer, Dennis and K{\"a}lvegren, Henrik and Haglund, Sebastian and Simioni, Edoardo and Bessa, Ricardo J and Wang, Yi},
  journal={International Journal of Forecasting},
  year={2025},
  publisher={Elsevier}
}

@article{mason1999boosting,
  title={Boosting algorithms as gradient descent},
  author={Mason, Llew and Baxter, Jonathan and Bartlett, Peter and Frean, Marcus},
  journal={Advances in neural information processing systems},
  volume={12},
  year={1999}
}

@inproceedings{sprangers2021probabilistic,
  title={Probabilistic gradient boosting machines for large-scale probabilistic regression},
  author={Sprangers, Olivier and Schelter, Sebastian and de Rijke, Maarten},
  booktitle={Proceedings of the 27th ACM SIGKDD conference on knowledge discovery \& data mining},
  pages={1510--1520},
  year={2021}
}

@article{hu2022conformalized,
  title={Conformalized temporal convolutional quantile regression networks for wind power interval forecasting},
  author={Hu, Jianming and Luo, Qingxi and Tang, Jingwei and Heng, Jiani and Deng, Yuwen},
  journal={Energy},
  volume={248},
  pages={123497},
  year={2022},
  publisher={Elsevier}
}

@article{wang2023conformal,
  title={Conformal asymmetric multi-quantile generative transformer for day-ahead wind power interval prediction},
  author={Wang, Wei and Feng, Bin and Huang, Gang and Guo, Chuangxin and Liao, Wenlong and Chen, Zhe},
  journal={Applied Energy},
  volume={333},
  pages={120634},
  year={2023},
  publisher={Elsevier}
}

@article{zhang2024two,
  title={Two-stage short-term wind power probabilistic prediction using natural gradient boosting combined with neural network},
  author={Zhang, Siyi and Liu, Mingbo and Xie, Min and Lin, Shunjiang},
  journal={Applied Soft Computing},
  volume={159},
  pages={111669},
  year={2024},
  publisher={Elsevier}
}

@article{li2020short,
  title={Short-term direct probability prediction model of wind power based on improved natural gradient boosting},
  author={Li, Yonggang and Wang, Yue and Wu, Binyuan},
  journal={Energies},
  volume={13},
  number={18},
  pages={4629},
  year={2020},
  publisher={MDPI}
}

@article{andrade2017improving,
  title={Improving renewable energy forecasting with a grid of numerical weather predictions},
  author={Andrade, Jos{\'e} R and Bessa, Ricardo J},
  journal={IEEE Transactions on Sustainable Energy},
  volume={8},
  number={4},
  pages={1571--1580},
  year={2017},
  publisher={IEEE}
}

@online{eliaopendata,
  title        = "Elia Open Data Platform",
  author       = {{Elia Transmission Belgium SA}},
  url = {https://opendata.elia.be/},
  urldate={2025-07-01},
}

@inproceedings{niayifar2015new,
  title={A new analytical model for wind farm power prediction},
  author={Niayifar, Amin and Port{\'e}-Agel, Fernando},
  booktitle={Journal of physics: conference series},
  volume={625},
  pages={012039},
  year={2015},
  organization={IOP Publishing}
}

@article{dhariwal2021diffusion,
  title={Diffusion models beat {GAN}s on image synthesis},
  author={Dhariwal, Prafulla and Nichol, Alexander},
  journal={Advances in neural information processing systems},
  volume={34},
  pages={8780--8794},
  year={2021}
}

@article{ho2022classifier,
  title={Classifier-free diffusion guidance},
  author={Ho, Jonathan and Salimans, Tim},
  journal={arXiv preprint arXiv:2207.12598},
  year={2022}
}

@inproceedings{rombach2022high,
  title={High-resolution image synthesis with latent diffusion models},
  author={Rombach, Robin and Blattmann, Andreas and Lorenz, Dominik and Esser, Patrick and Ommer, Bj{\"o}rn},
  booktitle={Proceedings of the IEEE/CVF conference on computer vision and pattern recognition},
  pages={10684--10695},
  year={2022}
}

@Article{Binsbergen2024b,
  author    = {Binsbergen, Diederik van and Daems, Pieter-Jan and Verstraeten, Timothy and Nejad, Amir and Helsen, Jan},
  journal   = {Journal of Physics: Conference Series},
  title     = {Scalable SCADA-Based Calibration for Analytical Wake Models Across an Offshore Cluster},
  year      = {2024},
  issn      = {1742-6596},
  month     = apr,
  number    = {1},
  pages     = {012014},
  volume    = {2745},
  doi       = {10.1088/1742-6596/2745/1/012014},
  publisher = {IOP Publishing},
}

@techreport{IEC61400-1,
  title        = {Wind energy generation systems — Part 1: Design requirements},
  type         = {International Standard},
  number       = {IEC 61400-1:2019},
  organisation = {International Electrotechnical Commission},
  edition      = {4},
  date         = {2019-02-08},
  year = {2019},
  isbn         = {9782832279724},
  url          = {https://webstore.iec.ch/publication/26423},
  urldate      = {2025-05-06}
}

@Article{Crespo1996,
  author    = {Crespo, A. and Hernandez, J.},
  journal   = {Journal of Wind Engineering and Industrial Aerodynamics},
  title     = {Turbulence characteristics in wind-turbine wakes},
  year      = {1996},
  issn      = {0167-6105},
  month     = jun,
  number    = {1},
  pages     = {71--85},
  volume    = {61},
  doi       = {10.1016/0167-6105(95)00033-x},
  publisher = {Elsevier BV},
}

@InProceedings{Katic1987,
  author     = {Ivan Kati{\'c} and J{\o}rgen H{\o}jstrup and Niels Jensen},
  title      = {A Simple Model for Cluster Efficiency},
  year       = {1987},
  comment    = {Katic (1987) improved the Jensen (1983) model by adding squared wake summation downwind. Furthermore, the model utilizes the ground effect of turbines by placing an imaginary turbine below the real turbine. Different Equations are described than in Jensen (1983), but they basically mean the same. They mention factors like ambient turbulence, turbine-induced turbulence and atmospheric stability highly influence the wake expansion factor.},
}

@misc{Pedersen2023,
  author    = {Pedersen, Mads M.},
  title     = {PyWake 2.5.0: An open-source wind farm simulation tool},
  year      = {2023},
  month     = {2},
  publisher = {DTU Wind, Technical University of Denmark},
  url       = {https://gitlab.windenergy.dtu.dk/TOPFARM/PyWake},
}

@Article{Binsbergen2024,
  author    = {van Binsbergen, Diederik and Daems, Pieter-Jan and Verstraeten, Timothy and Nejad, Amir R. and Helsen, Jan},
  journal   = {Wind Energy Science},
  title     = {Hyperparameter tuning framework for calibrating analytical wake models using SCADA data of an offshore wind farm},
  year      = {2024},
  issn      = {2366-7451},
  month     = jul,
  number    = {7},
  pages     = {1507--1526},
  volume    = {9},
  doi       = {10.5194/wes-9-1507-2024},
  publisher = {Copernicus GmbH},
}

@article{kato2023review,
  title={A review of nonconformity measures for conformal prediction in regression},
  author={Kato, Yuko and Tax, David MJ and Loog, Marco},
  journal={Conformal and Probabilistic Prediction with Applications},
  pages={369--383},
  year={2023},
  publisher={PMLR}
}

@article{ally2025modular,
  title={Modular deep learning approach for wind farm power forecasting and wake loss prediction},
  author={Ally, Stijn and Verstraeten, Timothy and Daems, Pieter-Jan and Now{\'e}, Ann and Helsen, Jan},
  journal={Wind Energy Science},
  volume={10},
  number={4},
  pages={779--812},
  year={2025},
  publisher={Copernicus Publications G{\"o}ttingen, Germany}
}

@article{steiner2017critical,
  title={Critical weather situations for renewable energies--Part A: Cyclone detection for wind power},
  author={Steiner, Andrea and K{\"o}hler, Carmen and Metzinger, Isabel and Braun, Axel and Zirkelbach, Mathias and Ernst, Dominique and Tran, Peter and Ritter, Bodo},
  journal={Renewable Energy},
  volume={101},
  pages={41--50},
  year={2017},
  publisher={Elsevier BV}
}

@techreport{smet2023probabilistic,
  title={Probabilistic storm forecasts for wind farms in the North Sea},
  author={Smet, Geert and Van den Bergh, Joris and Termonia, Piet},
  year={2023},
  institution={Copernicus Meetings}
}

@article{gneiting2007probabilistic,
  title={Probabilistic forecasts, calibration and sharpness},
  author={Gneiting, Tilmann and Balabdaoui, Fadoua and Raftery, Adrian E},
  journal={Journal of the Royal Statistical Society Series B: Statistical Methodology},
  volume={69},
  number={2},
  pages={243--268},
  year={2007},
  publisher={Oxford University Press}
}

@article{ahmadi2025enhancing,
  title={Enhancing Power Grid Stability with a Hybrid Framework for Wind Power Forecasting: Integrating Kalman Filtering, Deep Residual Learning, and Bidirectional {LSTM}},
  author={Ahmadi, Mehrnaz and Aly, Hamed and Khashei, Mehdi},
  journal={Energy},
  pages={137752},
  year={2025},
  publisher={Elsevier}
}

@article{weber2024intermittency,
  title={Intermittency or uncertainty? impacts of renewable energy in electricity markets},
  author={Weber, Paige and Woerman, Matt},
  journal={Journal of the Association of Environmental and Resource Economists},
  volume={11},
  number={6},
  pages={1351--1385},
  year={2024},
  publisher={The University of Chicago Press Chicago, IL}
}

@article{borisov2022deep,
  title={Deep neural networks and tabular data: A survey},
  author={Borisov, Vadim and Leemann, Tobias and Se{\ss}ler, Kathrin and Haug, Johannes and Pawelczyk, Martin and Kasneci, Gjergji},
  journal={IEEE transactions on neural networks and learning systems},
  volume={35},
  number={6},
  pages={7499--7519},
  year={2022},
  publisher={IEEE}
}

@article{grinsztajn2022tree,
  title={Why do tree-based models still outperform deep learning on typical tabular data?},
  author={Grinsztajn, L{\'e}o and Oyallon, Edouard and Varoquaux, Ga{\"e}l},
  journal={Advances in neural information processing systems},
  volume={35},
  pages={507--520},
  year={2022}
}

@book{corke2018wind,
  title={Wind energy design},
  author={Corke, Thomas C and Nelson, Robert C},
  year={2018},
  publisher={CRC Press}
}

@article{segalini2020blockage,
  title={Blockage effects in wind farms},
  author={Segalini, Antonio and Dahlberg, Jan-{\AA}ke},
  journal={Wind Energy},
  volume={23},
  number={2},
  pages={120--128},
  year={2020},
  publisher={Wiley Online Library}
}

\newpage
\onecolumn
\appendix

\subsection{Hyperparameter tuning}\label{app:hyp_opt}


The set of hyperparameters selected for model tuning is presented in Tables \ref{tab:tuning_xgboost} and \ref{tab:tuning_lightgbm}. These hyperparameters are described using the log-uniform distribution, denoted by $\log\mathcal{U}(a,b)$, and the discrete uniform distribution, denoted by $\mathcal{U}_D(a,b)$. It should be noted that the natural gradient boosting model\footnote{We utilize the natural gradient boosting implementation from the xgb-distribution package, as it has been shown to achieve superior computational efficiency compared to the ngboost library.} and the conformalized quantile regression model are implemented using XGBoost, whereas the Treeffuser model is based on LightGBM. Consequently, these models require a different set of hyperparameters to tune.


\begin{table}[htbp]
\centering
\caption{Tuning parameters for the NGBoost \& CQR (XGBoost) models.}
\label{tab:tuning_xgboost}
\begin{tabularx}{\textwidth}{XX}
    \toprule
    \textbf{Parameter} & \textbf{Values }\\
    \midrule
    eta              & $\mathcal{U}(0.01, 0.2)$ \\
    max\_depth        & $\mathcal{U}_D(3,7)$ \\
    min\_child\_weight & $\{1, 5, 10, 20\}$ \\
    gamma            & $\mathcal{U}(0, 1)$ \\
    subsample        & $\mathcal{U}(0.5, 1)$ \\
    \bottomrule
\end{tabularx}
\end{table}

\begin{table}[htbp]
\centering
\caption{Tuning parameters for the Treeffuser (LightGBM) model.}
\label{tab:tuning_lightgbm}
\begin{tabularx}{\textwidth}{XX}
    \toprule
     \textbf{Parameter} & \textbf{Values} \\
    \midrule
    n\_estimators          & $\mathcal{U}_D(100,3000)$ \\
    n\_repeats             & $\mathcal{U}_D(10,50)$ \\
    learning\_rate         & $\log\mathcal{U}(0.01, 1)$ \\
    early\_stopping\_rounds & $\mathcal{U}_D(10, 100)$ \\
    num\_leaves            & $\mathcal{U}_D(10, 100)$ \\
    \bottomrule
\end{tabularx}
\end{table}

\newpage

\subsection{Ensemble analysis}
\label{app:ensemble_analysis}

\begin{table*}[htbp]
\centering
\small
\setlength{\tabcolsep}{3pt} 
\caption{Per-farm breakdown of the experiments from Figure \ref{fig:mae_wf}.}
\begin{tabular}{lccccccc}
\toprule
 & ERA-5 & \makecell{DWD\\ICON-D2} & \makecell{DWD\\ICON-EU} & \makecell{ECMWF\\HRES} & \makecell{Met Office\\Global\\Hi-Res} & \makecell{MétéoFrance\\ARPEGE-EU} & Ensemble \\
\midrule
Belwind       & 6.7\% & 9.1\%  & 8.9\%  & 8.2\% & 8.5\%  & 9.5\%  & 7.0\% \\
Mermaid       & 8.3\% & 10.8\% & 10.6\% & 9.9\% & 10.2\% & 11.2\% & 9.0\% \\
Nobelwind     & 7.4\% & 10.1\% & 9.9\%  & 9.1\% & 9.6\%  & 10.7\% & 8.1\% \\
Norther       & 7.4\% & 10.1\% & 9.7\%  & 8.9\% & 9.7\%  & 10.2\% & 7.9\% \\
Northwester 2 & 8.4\% & 10.3\% & 10.1\% & 9.4\% & 9.9\%  & 10.6\% & 8.9\% \\
Northwind     & 7.4\% & 9.6\%  & 9.6\%  & 9.0\% & 9.2\%  & 10.3\% & 7.6\% \\
Rentel        & 8.3\% & 10.5\% & 10.3\% & 9.7\% & 10.2\% & 11.0\% & 8.5\% \\
Seastar       & 8.0\% & 10.3\% & 10.1\% & 9.4\% & 9.8\%  & 10.8\% & 8.4\% \\
C-Power       & 6.7\% & 9.3\%  & 9.0\%  & 8.2\% & 8.8\%  & 9.5\%  & 7.0\% \\
\bottomrule
\end{tabular}
\end{table*}

\begin{table*}[htbp]
\caption{Diebold-Mariano test output for single-provider forecasts versus ensemble model.}
\small
\centering
\begin{tabular}{lccc}
\toprule
Farm & NWP model & D-M statistic & P-value \\
\midrule
Belwind & DWD ICON-D2 & 16.7 & $8.26\times10^{-63}$ \\
Belwind & ECMWF HRES & 11.4 & $2.61\times10^{-30}$ \\
Belwind & DWD ICON-EU & 15.3 & $5.83\times10^{-53}$ \\
Belwind & MeteoFrance ARPEGE-EU & 18.2 & $1.17\times10^{-73}$ \\
Mermaid & DWD ICON-D2 & 15.6 & $5.41\times10^{-55}$ \\
Mermaid & ECMWF HRES & 9.8 & $1.55\times10^{-22}$ \\
Mermaid & DWD ICON-EU & 13.6 & $7.47\times10^{-42}$ \\
Mermaid & MeteoFrance ARPEGE-EU & 17.9 & $1.61\times10^{-71}$ \\
Nobelwind & DWD ICON-D2 & 15.6 & $4.40\times10^{-55}$ \\
Nobelwind & ECMWF HRES & 9.3 & $2.23\times10^{-20}$ \\
Nobelwind & DWD ICON-EU & 12.9 & $3.88\times10^{-38}$ \\
Nobelwind & MeteoFrance ARPEGE-EU & 17.9 & $2.25\times10^{-71}$ \\
Norther & DWD ICON-D2 & 16.3 & $2.00\times10^{-59}$ \\
Norther & ECMWF HRES & 9.7 & $2.38\times10^{-22}$ \\
Norther & DWD ICON-EU & 13.3 & $3.32\times10^{-40}$ \\
Norther & MeteoFrance ARPEGE-EU & 16.6 & $1.08\times10^{-61}$ \\
Northwester 2 & DWD ICON-D2 & 13.9 & $5.02\times10^{-44}$ \\
Northwester 2 & ECMWF HRES & 6.8 & $8.95\times10^{-12}$ \\
Northwester 2 & DWD ICON-EU & 13.0 & $1.80\times10^{-38}$ \\
Northwester 2 & MeteoFrance ARPEGE-EU & 15.3 & $1.25\times10^{-52}$ \\
Northwind & DWD ICON-D2 & 15.7 & $2.17\times10^{-55}$ \\
Northwind & ECMWF HRES & 12.9 & $4.55\times10^{-38}$ \\
Northwind & DWD ICON-EU & 14.9 & $4.65\times10^{-50}$ \\
Northwind & MeteoFrance ARPEGE-EU & 18.7 & $9.10\times10^{-78}$ \\
Rentel & DWD ICON-D2 & 14.7 & $8.14\times10^{-49}$ \\
Rentel & ECMWF HRES & 11.0 & $4.96\times10^{-28}$ \\
Rentel & DWD ICON-EU & 13.3 & $1.32\times10^{-40}$ \\
Rentel & MeteoFrance ARPEGE-EU & 17.8 & $1.03\times10^{-70}$ \\
Seastar & DWD ICON-D2 & 15.8 & $2.95\times10^{-56}$ \\
Seastar & ECMWF HRES & 10.6 & $4.59\times10^{-26}$ \\
Seastar & DWD ICON-EU & 13.5 & $9.89\times10^{-42}$ \\
Seastar & MeteoFrance ARPEGE-EU & 17.2 & $3.09\times10^{-66}$ \\
C-Power & DWD ICON-D2 & 17.7 & $7.31\times10^{-70}$ \\
C-Power & ECMWF HRES & 11.7 & $1.66\times10^{-31}$ \\
C-Power & DWD ICON-EU & 15.2 & $2.17\times10^{-52}$ \\
C-Power & MeteoFrance ARPEGE-EU & 18.8 & $7.64\times10^{-79}$ \\
\bottomrule
\end{tabular}
\end{table*}

\end{document}